\documentclass{acm_proc_article-sp}
\usepackage{url}
\usepackage{paralist}
\usepackage{colortbl}
\usepackage{graphicx}
\usepackage{subfigure}
\usepackage{booktabs} % For formal tables
\usepackage{adjustbox}
\usepackage{tikz}
\usepackage{amssymb}% http://ctan.org/pkg/amssymb
\usepackage{pifont}% http://ctan.org/pkg/pifont
\usepackage{adjustbox}
\usepackage{amsmath}
\usepackage{rotating}
\newcommand{\Datahub}{\texttt{Datahub}}
\newcommand{\openML}{\texttt{openML}}
\newcommand{\mlFlow}{\texttt{mlFlow}}
\newcommand{\seeDB}{\texttt{seeDB}}
\newcommand{\zenVisage}{\texttt{zenVisage}}
\newcommand{\QUDE}{\texttt{QUDE}}
\newcommand{\MLCube}{\texttt{MLCube}}
\newcommand{\GOODs}{\texttt{GOODs}}
\newcommand{\ProvDB}{\texttt{ProvDB}}
\newcommand{\Ground}{\texttt{Ground}}
\newcommand{\Rafiki}{\texttt{Rafiki}}

\newcommand{\TPOT}{\texttt{TPOT}}
\newcommand{\TransmogrifAI}{\texttt{TransmogrifAI}}
\newcommand{\DataXRay}{\texttt{Data X-Ray}}
\newcommand{\MacroBase}{\texttt{MacroBase}}
\newcommand{\DataHub}{\texttt{DataHub}}
\usepackage[ruled]{algorithm2e}

\SetAlFnt{\small}
\SetAlCapFnt{\small}
\SetAlCapNameFnt{\small}
\SetAlCapHSkip{0pt}
\IncMargin{-\parindent}

\usepackage{adjustbox}
\usepackage{array}
\usepackage{booktabs}
\usepackage{multirow}
\usepackage{rotating}

\graphicspath {{./} {Figures/}}
\newcolumntype{H}{>{\columncolor{black}\color{white}}c}

\begin{document}
\title{Automated Machine Learning: State-of-The-Art and Open Challenges}

\numberofauthors{3}
\author{
\alignauthor Radwa Elshawi\\
       \affaddr{University of Tartu, Estonia}\\ 
       \email{radwa.elshawi@ut.ee}\\
\alignauthor Mohamed Maher\\
       \affaddr{University of Tartu, Estonia}\\
       {\scriptsize \email{mohamed.abdelrahman\\@ut.ee}}
\alignauthor Sherif Sakr\\
       \affaddr{University of Tartu, Estonia}\\
       \email{sherif.sakr@ut.ee}
}

\maketitle
\begin{abstract}
Nowadays, machine learning techniques and algorithms are employed in almost every application domain (e.g., financial applications, advertising, recommendation systems, user behavior analytics). In practice, they are playing a crucial role in harnessing the power of massive amounts of data which we are currently producing every day in our digital world. In general, the process of building a high-quality machine learning model is an iterative, complex and time-consuming process that involves trying different algorithms and techniques in addition to having a good experience with effectively tuning their hyper-parameters. In particular, conducting this process efficiently requires solid knowledge and experience with the various techniques that can be employed. With the continuous and vast increase of the amount of data  in our digital world, it has been acknowledged that the number of knowledgeable data scientists can not scale to address these challenges. Thus, there was a crucial need for automating the process of building good machine learning models.
In the last few years, several techniques and frameworks have been introduced  to tackle the challenge of automating the process of \texttt{C}ombined  \texttt{A}lgorithm \texttt{S}election and \texttt{H}yper-parameter tuning (CASH) in the machine learning domain. The main aim of these techniques is to reduce the role of human in the loop and fill the gap for non-expert machine learning users by playing the role of the domain expert.

In this paper, we present a comprehensive survey for the state-of-the-art efforts in tackling the CASH problem. In addition, we highlight the research work of automating the other steps of the full complex machine learning pipeline (AutoML) from data understanding till model deployment. Furthermore, we provide a comprehensive coverage for the various tools and frameworks that have been introduced in this domain. Finally, we discuss some of the research directions and open challenges that need to be addressed in order to achieve the vision and goals of the AutoML process.
\end{abstract}

\vspace{0.5cm}
\section{Introduction}
Due to the increasing success of machine learning techniques in several application domains, they have been attracting a lot of attention from the research and business communities.
In general, the effectiveness of machine learning techniques mainly rests on the availability of massive datasets.
Recently, we have been witnessing a continuous exponential growth in the size of data produced
by various kinds of systems, devices and data sources. It has been reported that there are 2.5
quintillion bytes of data is being created everyday where 90\% of stored data in the world, has
been generated in the past two years only\footnote{Forbes: How Much Data Do We Create Every Day?
May 21, 2018}.
On the one hand, the more data that is available, the richer and the more robust the insights and the results that machine learning techniques can produce.
Thus, in the Big Data Era, we are witnessing many leaps achieved by machine and deep learning techniques in a wide range of fields~\cite{zomaya2017handbook,DBLP:reference/bdt/2019}.
On the other hand, this situation is raising a potential \emph{data science crisis}, similar to the software crisis~\cite{fitzgerald2012software}, due to the crucial need of having an increasing  number of data scientists with strong knowledge and good experience so that they are able to keep up with harnessing the power of the massive amounts of data which are produced daily. In particular, it has been acknowledged  that \emph{data scientists can not scale}\footnote{\url{https://hbr.org/2015/05/data-scientists-dont-scale}} and it is almost impossible to balance between the number of qualified data scientists and the required effort to manually analyze the increasingly growing sizes of available data. Thus, we are witnessing a growing focus and interest to support automating the process of building machine learning pipelines where the presence of a human in the loop can be dramatically reduced, or preferably eliminated.

\begin{figure*}
    \centering
		\includegraphics[width=\linewidth]{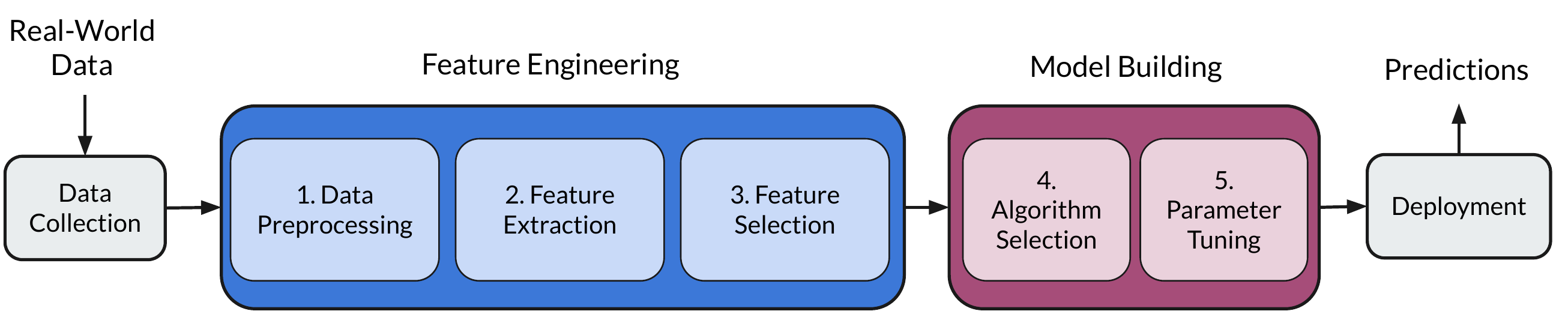}
		\caption{Typical Supervised Machine Learning Pipeline.}
		\label{fig:mlPipeline}
	\end{figure*}

In general, the process of building
a high-quality machine learning model is an iterative, complex and
time-consuming process that involves a number of steps (Figure~\ref{fig:mlPipeline}). In particular, a data scientist  is commonly \emph{challenged} with a large number of choices where informed decisions need to be taken. For example, the data scientist needs to select among a wide range of possible algorithms including classification or regression techniques (e.g.  Support Vector Machines, Neural Networks, Bayesian Models, Decision Trees, etc) in addition to tuning numerous hyper-parameters of the selected algorithm. In addition, the performance of the model can also be judged by various metrics (e.g., accuracy, sensitivity, specificity, F1-score).
Naturally, the decisions of the data scientist in each of these steps affect the performance and the quality of the developed  model~\cite{vafeiadis2015comparison,probst2017tune,pedregosa2011scikit}. For instance, in \texttt{yeast dataset}\footnote{\url{https://www.openml.org/d/40597}}, different parameter configurations of a Random Forest classifier result in different range of accuracy values, around 5\%\footnote{\url{https://www.openml.org/t/2073}}. Also, using different classifier learning algorithms leads to widely different performance values, around 20\% , for the fitted models   on the same dataset. Although making such decisions require solid knowledge and expertise, in practice, increasingly, users of machine learning tools are often non-experts who require \emph{off-the-shelf} solutions. Therefore, there has been a growing interest to \emph{automate} and \emph{democratize} the steps of building the machine learning pipelines.

In the last years, several techniques and frameworks have been introduced  to tackle the challenge of automating the process of Combined
Algorithm Selection and Hyper-parameter tuning (CASH) in the machine learning domain. These techniques have commonly formulated the problem  as an optimization problem that can be solved by  wide range of techniques~\cite{Kotthoff:2017:AAM:3122009.3122034,Feurer:2015:ERA:2969442.2969547,maher2019smartml}.  In general,
the $CASH$ problem is described as follows:

Given a set of machine learning algorithms $\mathbf{A}$ =
\{$A^{(1)}, A^{2},...$\}, and a dataset $D$ divided into disjoint training $D_{train}$, and
validation $D_{validation}$ sets. The goal is to find an algorithm $A^{(i)^{*}}$ where $A^{(i)}
\in \mathbf{A}$ and $A^{(i)^{*}}$ is a tuned version of $A^{(i)}$ that achieves the highest
generalization performance by training $A^{(i)}$ on $D_{train}$, and evaluating it on
$D_{validation}$. In particular, the goal of any CASH optimization technique is defined as:

\begin{figure}
\centering
	\includegraphics[width=\linewidth]{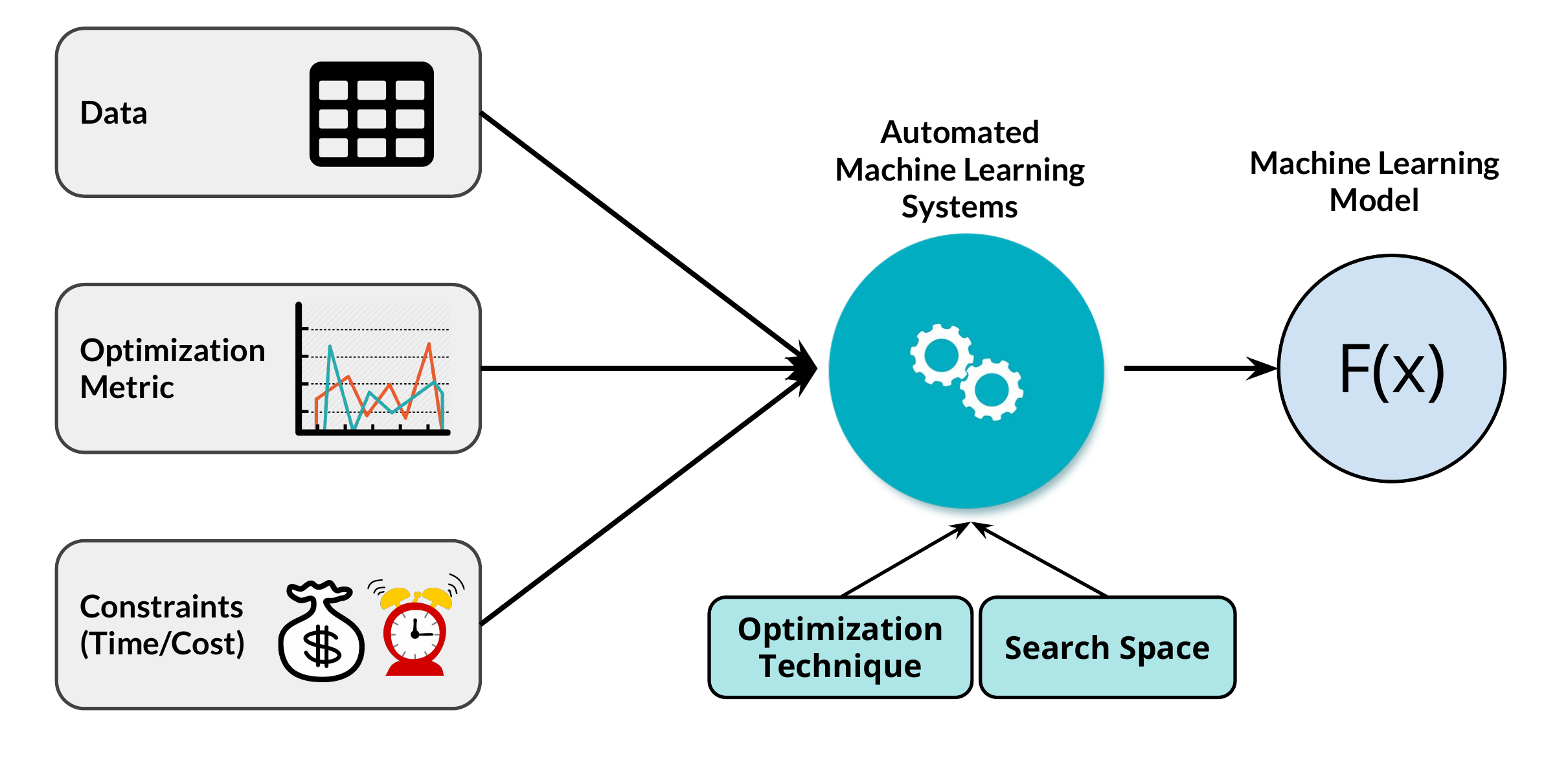}
	\caption{The general Workflow of the AutoML process.}
	\label{fig:Cash}
\end{figure}

\begin{center}
$A^{(i)^{*}} \in \underset{A~\epsilon~\mathbf{A}}{argmin} ~L(A^{(i)}, D_{train}, D_{validation})$
\end{center}

where L($A^{(i)}$, $D_{train}$, $D_{validation}$) is the loss function (e.g: error rate, false
positives, etc). In practice, one constraint for CASH optimization techniques is the \emph{time budget}. In particular, the aim of the optimization algorithm is to  select and tune a machine learning algorithm that can achieve (near)-optimal performance in terms of the user-defined evaluation metric (e.g., accuracy, sensitivity, specificity, F1-score) within the user-defined \emph{time budget} for the search process (Figure~\ref{fig:Cash}).

%\begin{figure*}
%	\includegraphics[width=\linewidth]{Figures/ModelingTaxonomy.jpg}
%	\caption{Building-blocks for AutoML framework to find the best pipeline/model for a machine %learning task {\color{red} \textbf{ToDo: To be removed and decomposed into small figures into %each section}}}
%	\label{Fig:ModelingTaxonomy}
%\end{figure*}

\vspace{0.4cm}
In this paper, we present a comprehensive
survey for the state-of-the-art efforts in tackling the CASH problem. In addition, we highlight the
research work of automating the other steps of the full end-to-end machine learning pipeline (AutoML) from
data understanding (pre-modeling) till model deployment (post-modeling)\footnote{We have prepared a repository with the state-of-the-art resources in the AutoML domain and made it available on \url{https://github.com/DataSystemsGroupUT/AutoML_Survey}}.
The remainder of this paper is organized as follows.
Section~\ref{Sec:Meta-learning} covers the various techniques that have been introduced to tackle the challenge of warm starting (meta-learning) for AutoML search problem in the context of machine learning and deep learning domains.
Section~\ref{Sec:NeuralArchitectureSearch} covers the various approaches that have been introduced for tackling the challenge of neural architecture search (NAS) in the context of deep learning.
Section~\ref{Sec:HyperparameterOptimization} focuses on the different approaches for automated hyper-parameter optimization.
Section~\ref{SEC:Framework} comprehensively covers the various tools and frameworks that have been implemented to tackle the CASH problem.
Section~\ref{SEC:Other} covers the state-of-the-art research efforts on tackling the automation aspects for the other building blocks (Pre-modeling and Post-Modeling)  of the complex machine learning pipeline. We discuss some of the research directions and open challenges that need to be addressed in order to achieve the vision and goals of the AutoML process in Section~\ref{SEC:open} before we finally conclude the paper in Section~\ref{SEC:Conclusion}

    \begin{figure*}
    \centering
		\includegraphics[width=0.7\linewidth]{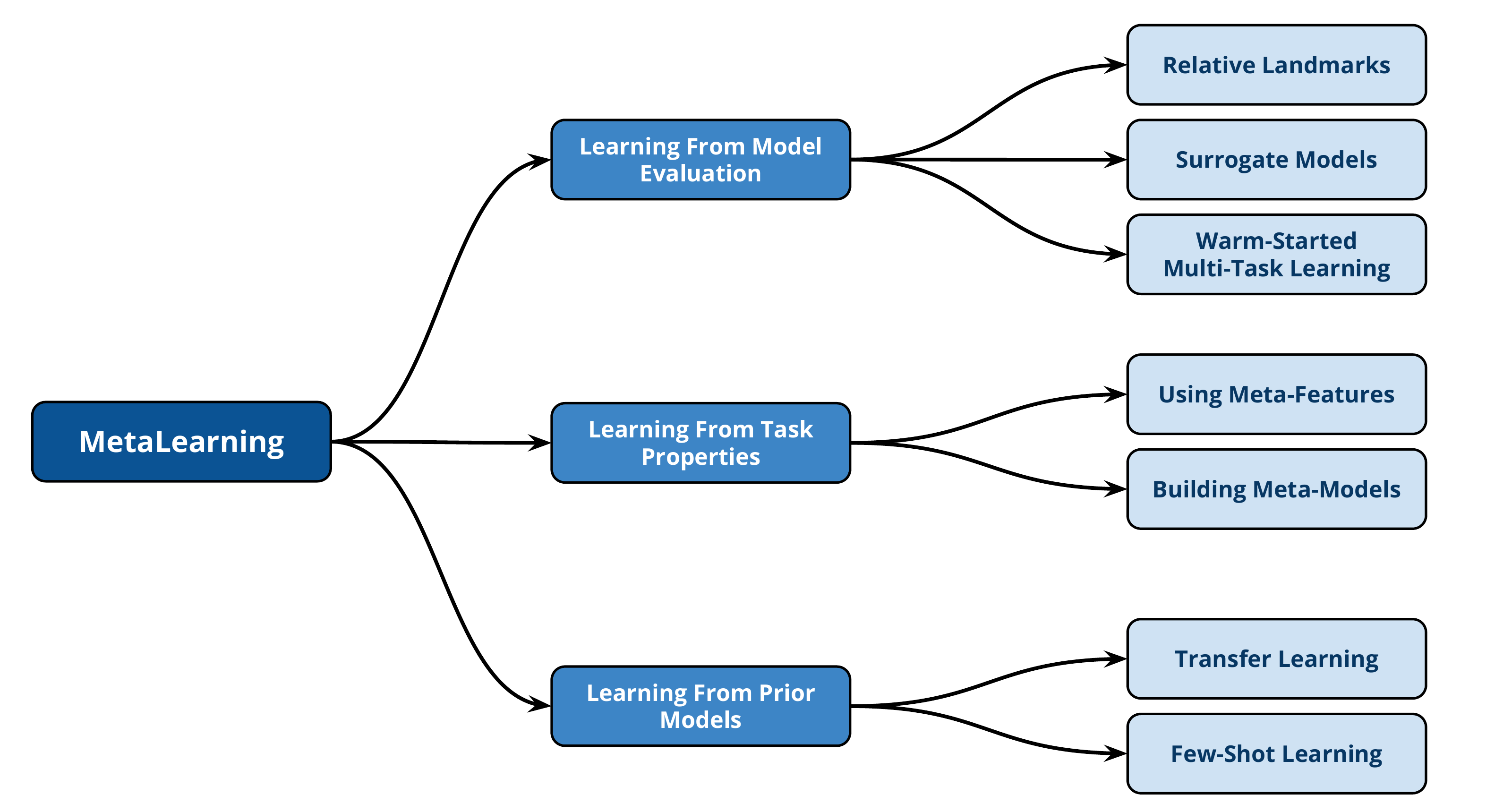}
		\caption{\textbf{A Taxonomy of Meta-Learning Techniques.}}
		\label{fig:metaLearning}
	\end{figure*}
	
\section{Meta-Learning}\label{Sec:Meta-learning}
In general, meta-learning can be described as the process of learning from previous experience gained during applying various learning algorithms on different kinds of data, and hence reducing the needed time to learn new tasks~\cite{brazdil2008metalearning}. In the context of machine learning, several \emph{meta learning}-techniques  have been introduced as an effective mechanism to tackle the challenge of warm start for optimization algorithms.  These techniques  can generally be categorized into three broad groups~\cite{DBLP:journals/corr/abs-1810-03548}: \emph{learning based on task properties}, \emph{learning from previous model evaluations} and \emph{learning from already pretrained models} (Figure~\ref{fig:metaLearning}).

One group of meta-learning techniques has been based on learning from task properties using the \emph{meta-features} that characterize a particular dataset~\cite{maher2019smartml}. Generally speaking, each prior task is characterized by a feature vector, of $k$ features, $m(t_j)$. Simply, information from a prior task $t_{j}$ can be transferred to a new task $t_{new}$ based on their similarity, where this similarity between $t_{new}$ and $t_{j}$ can be calculated based on the distance between their corresponding feature vectors. In addition, a meta learner $L$ can be trained on the feature vectors of prior tasks along with their evaluations $\textbf{P}$ to predict the performance of configurations $\theta_i$ on $t_{new}$.

Some of the commonly used meta features for describing datasets are simple meta features including number of instances, number of features, statistical features (e.g., skewness, kurtosis, correlation, co-variance, minimum, maximum, average), landmark features (e.g., performance of some landmark learning algorithms on a sample of the dataset), and information theoretic features (e.g., the entropy of class labels)~\cite{DBLP:journals/corr/abs-1810-03548}. In practice, the selection of the best set of meta features to be used is highly dependent on the application~\cite{bilalli2017predictive}. When computing the similarity between two tasks represented as two feature vectors of meta data, it is important to normalize these vectors or apply dimensionality reduction techniques such as principle component analysis~\cite{bilalli2017predictive,bardenet2013collaborative}. Another way to extract meta-features is to learn a joint distribution representation for set of tasks. One way to learn meta features is as follows:

 For a particular set of configurations $\theta_i$, evaluate all prior tasks $t_j$ and for each pair of configurations $\theta_a$ and $\theta_b$, generate a binary meta feature $m_{j,a,b}$ which is equal to one if the configuration $\theta_a$ outperforms the configuration $\theta_{b}$. The meta feature $m_{new,a,b}$ for a particular new task $t_{new}$ can be computed by learning meta-rules from every pair of configurations $\theta_a$ and $\theta_b$. Each of these learnt rules predicts whether a particular configuration $\theta_{a}$ outperforms configuration $\theta_b$ on prior tasks $t_{j}$, given their meta features.

Another way to learn from prior tasks properties is through building \emph{meta-models}. In this process, the aim is to build a meta model $L$ that learns complex relationships between meta features of prior tasks $t_j$. For a new task $t_{new}$, given the meta features for task $t_{new}$, model $L$ is used to recommend the best configurations. There exists a rich literature on using meta models for model configuration recommendations~\cite{soares2004meta,nisioti2018predicting,kopf2002combination,krizhevsky2012imagenet,giraud2008metalearning}. Meta models can also be used to rank a particular set of configurations by using the $K-$nearest neighbour model on the meta features of prior tasks and predicting the top $k$ tasks that are similar to new task $t_{new}$ and then ranking the best set of configurations of these similar tasks~\cite{brazdil2003ranking,dos2004selection}. Moreover, they can also be used to predict the performance of new task based on a particular configuration~\cite{reif2014automatic,guerra2008predicting}. This gives an indication about how good or bad this configuration can be, and whether it is worth evaluating it on a particular new task.

Another group of meta-learning techniques are based on \emph{learning from previous model evaluation}. In this context, the problem is formally defined as follows.

Given a set of machine learning tasks $t_j\in T$, their corresponding learned models along their hyper-parameters $\theta\in \Theta$ and $P_{i,j}=P(\theta_i,t_j)$, the problem is to learn a meta-learner $L$ that is trained on meta-data $\textbf{P}\cup \textbf{P}_{new}$ to predict recommended configuration $\Theta_{new}^{*}$ for a new task $t_{new}$, where $T$ is the set of all prior machine learning tasks. $\Theta$ is the configuration space (hyper-parameter setting, pipeline components, network architecture, and network hyper-parameter), $\Theta_{new}$ is the configuration space for a new machine learning task $t_{new}$, $\textbf{P}$ is the set of all prior evaluations $P_{i,j}$ of configuration $\theta_{i}$ on a prior task $t_j$, and $\textbf{P}_{new}$ is a set of evaluations $P_{i,new}$ for a new task $t_{new}$.

One way to get hyper-parameter recommendations for a new task $t_{new}$ is to recommend based on similar prior tasks $t_j$. \iffalse The notations above are used consistently throughout this paper.\fi In the following, we will go through three different ways: \emph{relative landmarks}, \emph{surrogate models} and \emph{Warm-Started Multi-task Learning} for relating similarity between $t_{new}$ and $t_j$. One way to measure the similarity between $t_{new}$ and $t_j$ is using the \emph{relative landmarks} that measures the performance difference between two model configurations on the same task~\cite{furnkranz2001evaluation}. Two tasks $t_{new}$ and $t_{j}$ are considered similar if their relative landmarks performance of the considered configurations are also similar. Once similar tasks have been identified, a meta learner can be trained on the evaluations $P_{i,j}$ and $P_{i,new}$ to recommend new configurations for task $t_{new}$. Another way to define learning from model evaluations is through \emph{surrogate models}~\cite{wistuba2018scalable}. In particular, \emph{Surrogate models} get trained on all prior evaluations of for all prior tasks $t_{j}$. Simply, for a particular task $t_j$, if the surrogate model can predict accurate configuration for a new task $t_new$, then tasks $t_new$ and $t_j$ are considered similar. \emph{Warm-Started Multi-task Learning} is another way to capture similarity between $t_j$ and $t_{new}$. Warm-Started Multi-task Learning uses the set of prior evaluations $\textbf{P}$ to learn a joint task representation~\cite{perrone2017multiple} which is used to train surrogate models on prior tasks $t_j$ and integrate them to a feed-forward neural network to learn a joint task representation that can predict accurately a set of evaluation $t_{new}$.

Learning from prior models can be done using \emph{Transfer learning}~\cite{pan2010survey}, which is the process of utilization of pretrained models on prior tasks $t_j$  to be adapted on a new task $t_{new}$, where tasks $t_j$ and $t_{new}$ are similar. Transfer learning has received lots of attention especially in the area of neural network. In particular, neural network architecture and neural network parameters are trained on prior task $t_j$ that can be used as an initialization for model adaptation on a new task $t_{new}$. Then, the model can be fine-tuned~\cite{bengio2012deep,baxter1995learning,caruana1995learning}. It has been shown that neural networks trained on big image datasets such as ImageNet~\cite{krizhevsky2012imagenet} can be transferred as well to new tasks~\cite{sharif2014cnn,donahue2014decaf}. Transfer learning usually works well when the new task to be learned is similar to the prior tasks, otherwise transfer learning may lead to unsatisfactory results~\cite{yosinski2014transferable}. In addition, prior models can be used in \emph{Few-Shot Learning} where a model is required to be trained using a few training instances given the prior experience gained from already trained models on similar tasks. For instance, learning a common feature representation for different tasks can be used to take the advantage of utilizing pretrained models that can initially guide the model parameters optimization for the few instances available. Some attempts have tackled the few-shot learning. For example, Snell et al.~\cite{snell2017prototypical} presented an approach where prototypical networks were used to map instances from separate tasks into a similar dimensional space to use the same model. Moreover, Ravi and Larochelle \cite{ravi2016optimization} proposed an LSTM network as a meta-learner that is trained to learn the update rule for fitting the neural network learner. For instance, the loss and gradient are passed from the learner to the meta-learner networks which in turn updates them before modifying the learner parameters. Mishra et al.~\cite{mishra2017simple} proposed a meta-learner that tries to learn a common feature vector among different tasks. The architecture of this meta-learner consists of an architecture of convolution layers in addition to attention layers that tries to learn useful parts from tasks that can be used to make it more generic for new tasks.

\section{Neural Architecture Search for Deep Learning}\label{Sec:NeuralArchitectureSearch}

\begin{figure}
		\includegraphics[width=\linewidth]{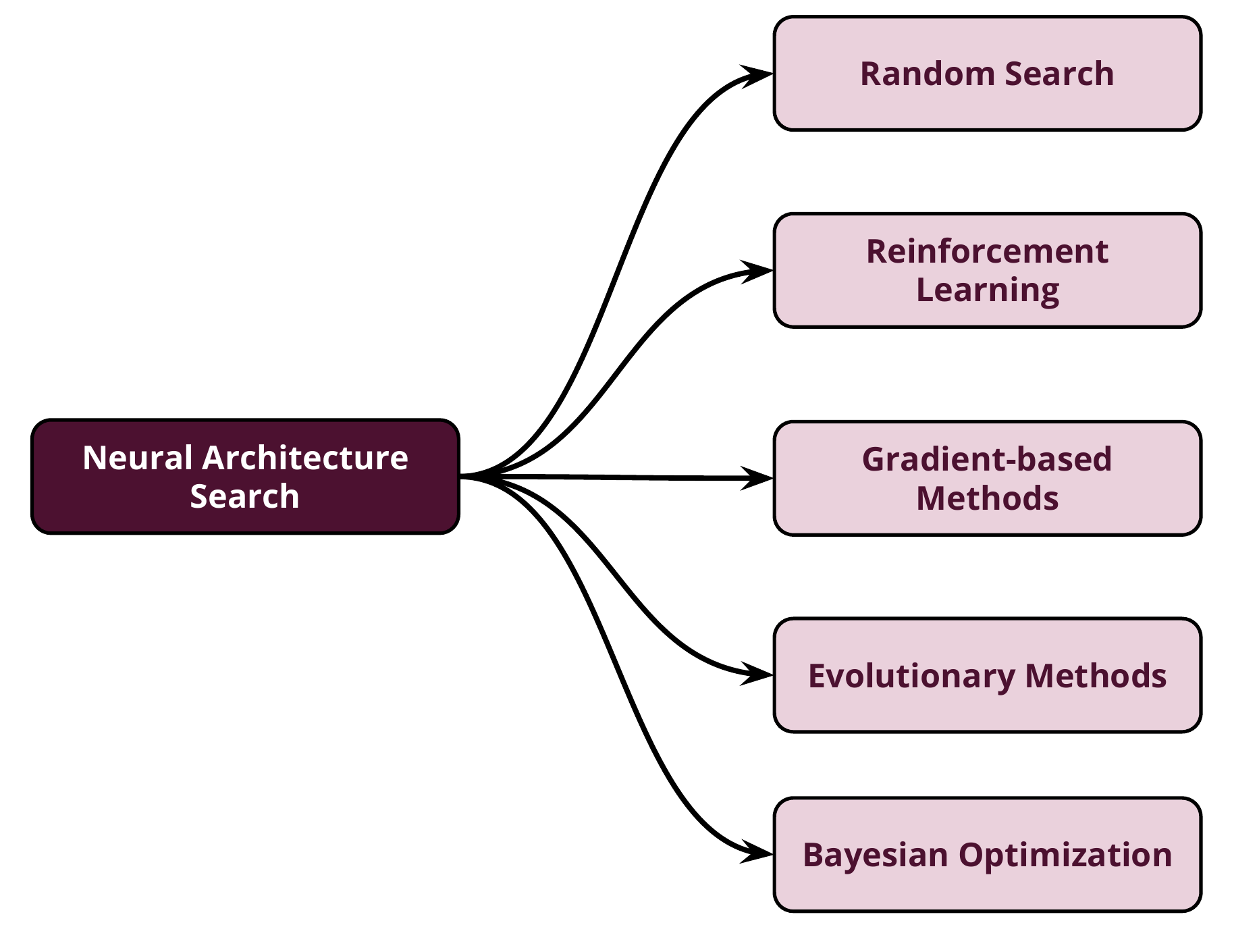}
		\caption{\textbf{A Taxonomy for the  Neural Network Architecture Search (NAS) Techniques }}
		\label{fig:nas}
	\end{figure}

In general, deep learning techniques represent a subset of machine learning methodologies that are based on artificial neural networks (ANN) which are mainly inspired by the neuron structure of the human brain~\cite{bengio2009learning}.
It is described as \emph{deep} because it has more than one layer of nonlinear feature transformation. Neural Architecture Search (NAS) is a fundamental step in automating the machine learning process and has been successfully used to design the model architecture for image and language tasks~\cite{zoph2016neural,zoph2018learning,cai2018efficient,liu2018progressive,liu2017hierarchical}. Broadly, NAS techniques falls into five main categories including \emph{random search}, \emph{reinforcement learning}, \emph{gradient-based methods}, \emph{evolutionary methods}, and \emph{Bayesian optimization} (Figure~\ref{fig:nas}).

\emph{Random search} is one of the most naive and simplest approaches for network architecture search. For example, Hoffer el al.~\cite{Hoffer:2017:TLG:3294771.3294936} have presented an approach to find good network architecture using random search combined with well-trained set of shared weights. Li et Talwalkar~\cite{li2019random} proposed new network architecture search baselines that are based on random search with early-stopping for hyper-parameter optimization. Results show that random search along with early-stopping achieves the state-of-the-art network architecture search results on two standard NAS bookmarkers which are PTB and CIFAR-10 datasets.

\emph{Reinforcement learning}~\cite{sutton1998introduction} is another approach that has been used to find the best network architecture. Zoph and Le~\cite{zoph2016neural} used a recurrent neural network (LSTM) with reinforcement to compose neural network architecture. More specifically, recurrent neural network is trained through a gradient based search algorithm called \texttt{REINFORCE}~\cite{williams1992simple} to maximize the expected accuracy of the generated neural network architecture. Baker et al.~\cite{baker2016designing} introduced a meta-modeling algorithm called \texttt{MetaQNN} based on reinforcement learning to automatically generate the architecture of convolutional neural network for a new task. The convolutional neural network layers are chosen sequentially by a learning agent that is trained using $Q-$learning with $\epsilon-$greedy exploration technique. Simply, the agent explores a finite search space of a set of architectures and iteratively figures out architecture designs with improved performance on the new task to be learnt.

\emph{Gradient-based optimization} is another common way for neural network architecture search. Liu et al.~\cite{liu2018darts} proposed an approach based on continuous relaxation of the neural architecture allowing using a gradient descent for architecture search. Experiments showed that this approach excels in finding high-performance convolutional architectures for image classification tasks on CIFAR-10, and ImageNet datasets. Shin et al.~\cite{shin2018differentiable} proposed a gradient-based optimization approach for learning the network architecture and parameters simultaneously. Ahmed and Torresani~\cite{Ahmed_2018} used gradient based approach to learn network architecture. Experimental results on two different networks architecture ResNet and ResNeXt show that this approach yields to better accuracy and significant reduction in the number of parameters.

Another direction for architecture search is \emph{evolutionary algorithms} which are well suited for optimizing arbitrary structure. Miller et al.~\cite{miller1989designing} considered an evolutionary algorithm to propose the architecture of the neural network and network weights as well. Many evolutionary approaches based on genetic algorithms are used to optimize the neural networks architecture and weights~\cite{stanley2002evolving,stanley2009hypercube,angeline1994evolutionary} while others rely on hierarchical evolution~\cite{liu2017hierarchical}. Some recent approaches consider using the multi-objective evolutionary architecture search to optimize training time, complexity and performance~\cite{lu2018nsga,elsken2018efficient} of the network. \texttt{LEAF}~\cite{liang2019evolutionary} is an evolutionary AutoML framework that optimizes hyper-parameters, network architecture and the size of the network. LEAF uses \texttt{CoDeepNEAT}~\cite{miikkulainen2019evolving} which is a powerful evolutionary algorithm based on \texttt{NEAT}~\cite{real2017large}. LEAF achieved the state-of-the-art performance results on medical image classification and natural language analysis. For supervised learning tasks, evolutionary based approaches tend to outperform reinforcement learning approaches especially  when the neural network architecture is very complex due to having millions of parameters to be tuned. For example, the best performance achieved on ImageNet and CIFAR-10 has been obtained using evolutionary techniques~\cite{real2018regularized}.

\emph{Bayesian optimization}  based on Gaussian processes has been used  by Kandasamy et al.~\cite{k2018neural} and Swersky et al.~\cite{swersky2014raiders} for tackling the neural architecture search problem. In addition, lots of work focused on using tree based models such as random forests and tree Parzen estimators~\cite{bergstra2011algorithms} to effectively optimize the network architecture as well as its hyper-parameters~\cite{bergstra2013making,domhan2015speeding,mendoza2016towards}. Bayesian optimization may outperform evolutionary algorithms in some problems as well~\cite{klein2018towards}.

\section{Hyper-parameter Optimization}\label{Sec:HyperparameterOptimization}
    \begin{figure*}
		\includegraphics[width=\linewidth]{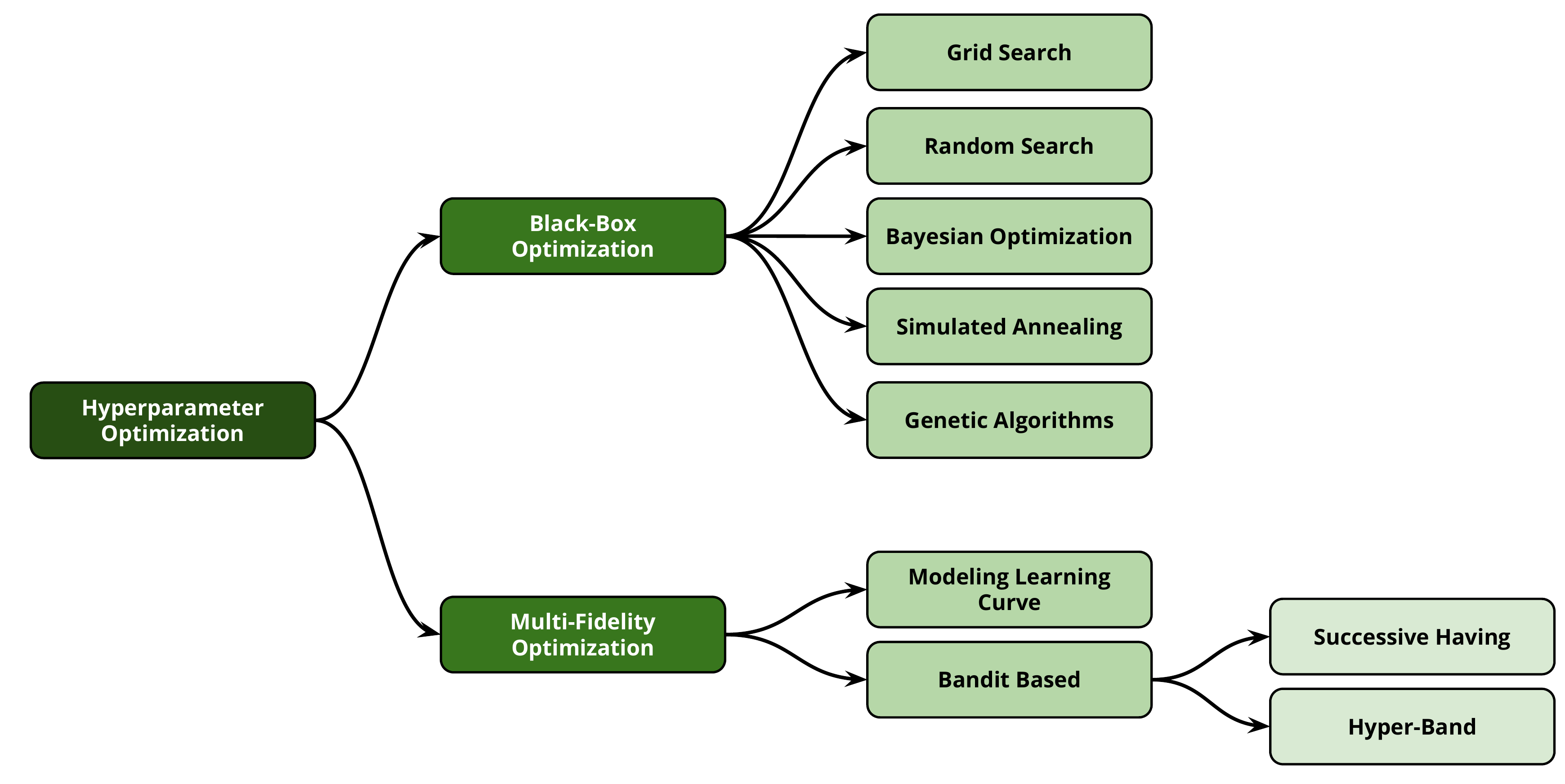}
		\caption{\textbf{A Taxonomy for the Hyper-parameter Optimization Techniques.}}
		\label{fig:hpo}
	\end{figure*}

After choosing the model pipeline algorithm(s) with the highest potential for achieving the top performance on the input dataset, the next step is tuning the hyper-parameters of the model in order to further optimize the model performance. It is worth mentioning that some tools have democratized the space of different learning algorithms in discrete number of model pipelines \cite{Kotthoff:2017:AAM:3122009.3122034,Feurer:2015:ERA:2969442.2969547}. So, the model selection itself can be considered as a categorical parameter that needs to be tuned in the first place before modifying its hyper-parameters. In general, several hyper-parameter optimization techniques have been based and borrowed ideas from the domains of statistical model selection and traditional optimization techniques~\cite{davis1991handbook,pelikan1999boa,polak2012optimization}. In principle, the automated hyper-parameter tuning techniques can be classified into two main categories: \emph{black-box optimization
techniques} and  \emph{multi-fidelity optimization techniques} (Figure~\ref{fig:hpo}).

\subsection{Black-Box optimization}\label{Sec:Black-Boxoptimization}

\emph{Grid search} is a simple basic solution for the hyper-parameter optimization~\cite{montgomery2017design} in which all combinations of hyper-parameters are evaluated.
Thus, grid search is computationally expensive, infeasible and suffers from the \emph{curse of dimensionality} as the number of trails grows exponentially with the number of hyper-parameters~\cite{bellman2015adaptive}. Another alternative is \emph{random search} in which it samples configurations at random until a particular budget $B$ is exhausted~\cite{bergstra2012random}. Given a particular computational budget $B$, random search tends to find better solutions than grid search~\cite{montgomery2017design}. One of the main advantages of random search, and grid search is that they can be easily parallelized over a number of workers which is essential when dealing with big data.

\emph{Bayesian Optimization} is one of the state-of-the-art black-box optimization techniques which is tailored for expensive objective functions~\cite{kushner1964new,zhilinskas1975single,mockus1978application,jones1998efficient}.
Bayesian optimization has received huge attention from the machine learning community in tuning deep neural networks for different tasks including
classification tasks~\cite{snoek2015scalable,snoek2012practical}, speech
recognition~\cite{dahl2013improving} and natural language processing~\cite{melis2017state}.
Bayesian optimization consists of two main components which are surrogate models for modeling the objective function and an acquisition function that measures the value that would be generated by the evaluation of the objective function at a new point. Gaussian processes have become the standard surrogate for modeling the objective function in
Bayesian optimization~\cite{snoek2012practical,martinez2014bayesopt}. One of the main limitations
of the Gaussian processes is the cubic complexity to the number of data points which limits their
parallelization capability. Another limitation is the poor scalability when using the standard
kernels. Random forests ~\cite{breiman2001random} are another choice for modeling the objective
function in Bayesian optimization. First, the algorithm starts with growing $B$ regression trees, each of which is built using $n$ randomly selected data points with replacement from training data of size $n$. For each tree, a split node is chosen from $d$ algorithm parameters. The minimum number of points are considered for further split are set to 10 and the number of trees $B$ to grow is set be $10$ to maintain low computational overhead. Then, the random forest predicted mean and variance for each new configuration is computed. The random forests' complexity of the fitting and predicting variances are $O(n\log n)$ and $O(\log n)$ respectively which is much better compared to the Gaussian process. Random forests are used by the Sequential Model-based Algorithm Configuration (SMAC) library~\cite{hutter2011sequential}. In general Tree-structured Parzen Estimator (TPE)~\cite{bergstra2011algorithms} does not define a predictive
distribution over the objective function but it creates two density functions that act as
generative models for all domain variables. Given a percentile $\alpha$, the observations are
partitioned into two sets of observations (good observations and bad observations) where simple
Parzen windows are used to model the two sets. The ratio between the two density functions
reflects the expected improvement in the acquisition function and is used to recommend new
configurations for hyper-parameters. Tree-Structured Parzen estimator (TPE) has shown great performance for hyper-parameter
optimization
tasks~\cite{bergstra2011algorithms,bergstra2013making,eggensperger2013towards,falkner2018bohb,sparks2015automating}.

\emph{Simulated Annealing} is a hyper-parameter optimization approach which is inspired by the
metallurgy technique of heating and controlled cooling of
materials~\cite{kirkpatrick1983optimization}. This optimization techniques goes through a number of steps. First, it randomly chooses a single value (current state) to be applied to all hyper-parameters and then evaluates model performance based on it. Second, it randomly updates the value of one of the hyper-parameters by picking a value from the immediate neighborhood to get neighboring state. Third, it evaluates the model performance based on the
neighboring state. Forth, it compares the performance obtained from the current and neighbouring
states. Then, the user chooses to reject or accept the neighbouring state as a current state based on some criteria.

\emph{Genetic Algorithms} (GA) are inspired by the process of natural selection~\cite{holland1992adaptation}. The main idea of genetic-based optimization techniques is simply applying multiple genetic operations to a population of configurations. For example, the \emph{crossover} operation simply takes two parent \emph{chromosomes} (configurations) and combines their genetic information to generate new \emph{offspring}. More specifically, the two parents configurations are cut at the same crossover point. Then, the sub-parts to the right of that point are swapped between the two parents chromosomes. This contributes to two new \emph{offspring} (child configuration). Mutation randomly chooses a chromosome and mutates one or more of its parameters that results in totally new chromosome.

\subsection{Multi-fidelity optimization}\label{Sec:Multifidelityoptimization}
Multi-fidelity optimization is an optimization technique which focuses on decreasing the
evaluation cost by combining a large number of cheap low-fidelity evaluations and a small number
of expensive high-fidelity evaluation~\cite{fernandez2016review,march2012provably,hu2019multi}.
In practice, such optimization technique is essential when dealing with big datasets as training one
hyper-parameter may take days. More specifically, in multi-fidelity optimization, we can evaluate
samples in different levels. For example, we may have two evaluation functions: \emph{high-fidelity}
evaluation and \emph{low-fidelity} evaluation. The high-fidelity evaluation outputs precise evaluation
from the whole dataset. On the other hand, the low-fidelity evaluation is a cheaper evaluation from a subset of the dataset. The idea behind the multi-fidelity evaluation is to use many low-fidelity evaluation to reduce the total evaluation cost. Although the low fidelity optimization results in cheaper evaluation cost that may suffer from optimization performance, but the speedup achieved is more significant than the approximation error.

\emph{Modeling learning curves} is an optimization technique that models learning curves during
hyper-parameter optimization and decides whether to allocate more resources or to stop the training
procedure for a particular configuration. For example, a curve may model the
performance of a particular hyper-parameter on an increasing subset of the dataset. Learning curve
extrapolation is used in predicting early termination for a particular
configuration~\cite{jones1998efficient}; the learning process is terminated if the performance of the
predicted configuration is less than the performance of the best model trained so far in the
optimization process. Combining early predictive termination criterion with Bayesian optimization
leads to more reduction in the model error rate than the vanilla Bayesian black-box optimization. In addition, such technique resulted in speeding-up the optimization by a factor of 2 and achieved the state-of-the-art neural network on CIFAR-10 dataset~\cite{domhan2015speeding}.

\emph{Bandit-based algorithms} have shown to be powerful in tackling deep learning optimization challenges. In the following, we consider two strategies of the bandit-based techniques which are the \emph{Successive halving} and \emph{HyperBand}.
\emph{Successive halving} is a bandit-based powerful multi-fidelity technique in which given a budget $B$, first, all the configurations are evaluated. Next, they are ranked based on their performance. Then, half of these configurations that performed worse than the others are removed. Finally, the budget of
the previous steps are doubled and repeated until only one algorithm remains. It is shown that the successive halving outperforms the uniform budget allocation technique in terms of the computation time, and the number of iterations required~\cite{jamieson2016non}. On the other hand, successive halving suffer from the following problem. Given a time budget $B$, the user has to choose, in advance, whether to consume the larger portion of the budget exploring a large number of configurations while spending a small portion of the time budget on tuning each of them or to consume the large portion of the budget on exploring few configurations while spending the larger portion of the budget on tuning them.

\emph{HyperBand} is another bandit-based powerful multi-fidelity hedging technique that optimizes the search space when selecting from randomly sampled configurations~\cite{li2016hyperband}. More specifically, partition a given budget $B$ into combinations of number of configurations and budget assigned to each configuration. Then, call successive halving technique on each random sample configuration.
Hyper-Band shows great success with deep neural networks and perform better than random search and Bayesian optimization.

\section{Tools and Frameworks}
\label{SEC:Framework}
In this section, we provide a comprehensive overview of several tools and frameworks that have been implemented to automate the process of combined algorithm selection and hyper-parameter optimization process. In general, these tools and frameworks can be classified into three main categories: \emph{centralized}, \emph{distributed}, and \emph{cloud-based}.

\subsection{Centralized Frameworks}
Several tools have been implemented on top of widely used \emph{centralized} machine learning packages which are designed to run in a \emph{single} node (machine). In general, these tools are suitable for handling small and medium sized datasets. For example, \emph{Auto-Weka}\footnote{\url{https://www.cs.ubc.ca/labs/beta/Projects/autoweka/}} is considered as the first and pioneer machine learning automation framework~\cite{Kotthoff:2017:AAM:3122009.3122034}. It was implemented in Java on top of \texttt{Weka}\footnote{\url{https://www.cs.waikato.ac.nz/ml/weka/}}, a popular machine learning library that has a wide range of machine learning algorithms. \texttt{Auto-Weka} applies Bayesian optimization using Sequential Model-based Algorithm Configuration (\texttt{SMAC})~\cite{hutter2011sequential} and
tree-structured parzen estimator (TPE) for both algorithm selection and hyper-parameter optimization (Auto-Weka uses SMAC as its default optimization algorithm but the user can configure the tool to use TPE). In particular, SMAC tries to draw the relation between algorithm performance and a given set of hyper-parameters by estimating the predictive mean and variance of their performance along the
trees of a random forest model. The main advantage of using SMAC is its robustness by having the ability to discard low performance parameter configurations quickly after the evaluation on low number of dataset folds. SMAC shows better performance on experimental results compared to TPE~\cite{hutter2011sequential}.

\emph{$Auto-MEKA_{GGP}$}~\cite{de2018automated} focuses on the AutoML task for multi-label classification problem~\cite{Tsoumakas10miningmulti-label} that aims to learn models from data capable of representing the relationships between input attributes and a set of class labels, where each instance may belong to more than one class. Multi-label classification has lots of applications especially in medical diagnosis in which a patient may be diagnosed with more than one disease. $Auto-MEKA_{GGP}$ is a grammar-based genetic programming framework that can handle complex multi-label classification search space and simply explores the hierarchical structure of the problem. $Auto-MEKA_{GGP}$ takes as input both of the dataset and a grammar describing the hierarchical search space of the hyper-parameters and the learning algorithms from \texttt{MEKA}\footnote{\url{http://waikato.github.io/meka/}} framework~\cite{read2016meka}. $Auto-MEKA_{GGP}$ starts by creating an initial set of trees representing the multi-label classification algorithms by randomly choosing valid rules from the grammar, followed by the generation of derivation trees. Next, map each derivation tree to a specific multi-label classification algorithm. The initial trees are evaluated on the input dataset by running the learning algorithm, they represent, using MEKA framework. The quality of the individuals are assessed using different measures such as fitness function. If a stopping condition is satisfied (e.g. a quality criteria ), a set of individuals (trees) are selected in a tournament selection. Crossover and mutation are applied in a way that respect the grammar constraints on the selected individuals to create a new population. At the end of the evolution, the best set of individuals representing the well performing set of multi-label tuned classifiers are returned.

\emph{Auto-Sklearn}\footnote{\url{https://github.com/automl/auto-sklearn}} has been implemented on top of \texttt{Scikit-Learn}\footnote{\url{https://scikit-learn.org/}}, a popular
Python machine learning package~\cite{Feurer:2015:ERA:2969442.2969547}. \texttt{Auto-Sklearn} introduced the idea of meta-learning in the initialization of combined algorithm selection and
hyper-parameter tuning. It used SMAC as a Bayesian optimization technique too. In
addition, ensemble methods were used to improve the performance of output models. Both meta-learning and ensemble methods improved the performance of \emph{vanilla} SMAC optimization. \emph{hyperopt-Sklearn}~\cite{komer2014hyperopt} is another AutoML framework which is based on Scikit-learn machine learning library. Hyperopt-Sklearn uses \texttt{Hyperopt}~\cite{bergstra2013hyperopt} to define the search space over the possible Scikit-Learn main components including the learning and preprocessing algorithms. Hyperpot supports different optimization techniques including random search, and different Bayesian optimizations for exploring the search spaces which are characterized by different types of variables including categorical, ordinal and continuous.

\emph{TPOT}\footnote{\url{https://automl.info/tpot/}} framework represents another type of solutions that has been implemented on top of \texttt{Scikit-Learn}~\cite{pmlr-v64-olson_tpot_2016}. It is based on genetic programming by exploring many different possible pipelines of feature engineering and learning algorithms. Then, it finds the best one out of them. \emph{Recipe}~\cite{DBLP:conf/eurogp/SaPOP17} follows the same optimization
procedure as \TPOT~ using genetic programming, which in turn exploits the advantages of a global search. However, it considers the unconstrained search problem in \TPOT, where resources can be spent into generating and evaluating invalid solutions by adding a grammar that avoids the generation of invalid pipelines, and can speed up optimization process. Second, it works with a bigger search space of different model configurations than \emph{Auto-SkLearn} and \TPOT.

\emph{ML-Plan}\footnote{\url{https://github.com/fmohr/ML-Plan}} has been proposed to tackle the
composability challenge on building machine learning pipelines~\cite{DBLP:journals/ml/MohrWH18}. In particular, it integrates a super-set of both \texttt{Weka} and \texttt{Scikit-Learn} algorithms to construct a full pipeline. ML-Plan tackles the challenge of the search problem for finding optimal machine learning pipeline using hierarchical task network algorithm where the search space is modeled as a large tree graph where each leaf node is considered as a goal node of a full pipeline. The graph traversal starts from the root node to one of the leaves by selecting some random paths.
The quality of a certain node in this graph is measured after making $n$ such random complete traversals and taking the minimum as an estimate for the best possible solution that can be found. The initial results of this approach has shown that the composable pipelines over \texttt{Weka} and \texttt{Scikit-Learn} does not significantly outperform the outcomes from \texttt{Auto-Weka} and \texttt{Auto-Sklearn} frameworks because it has to deal  with larger search space.

\begin{figure*}
		\includegraphics[width=\linewidth]{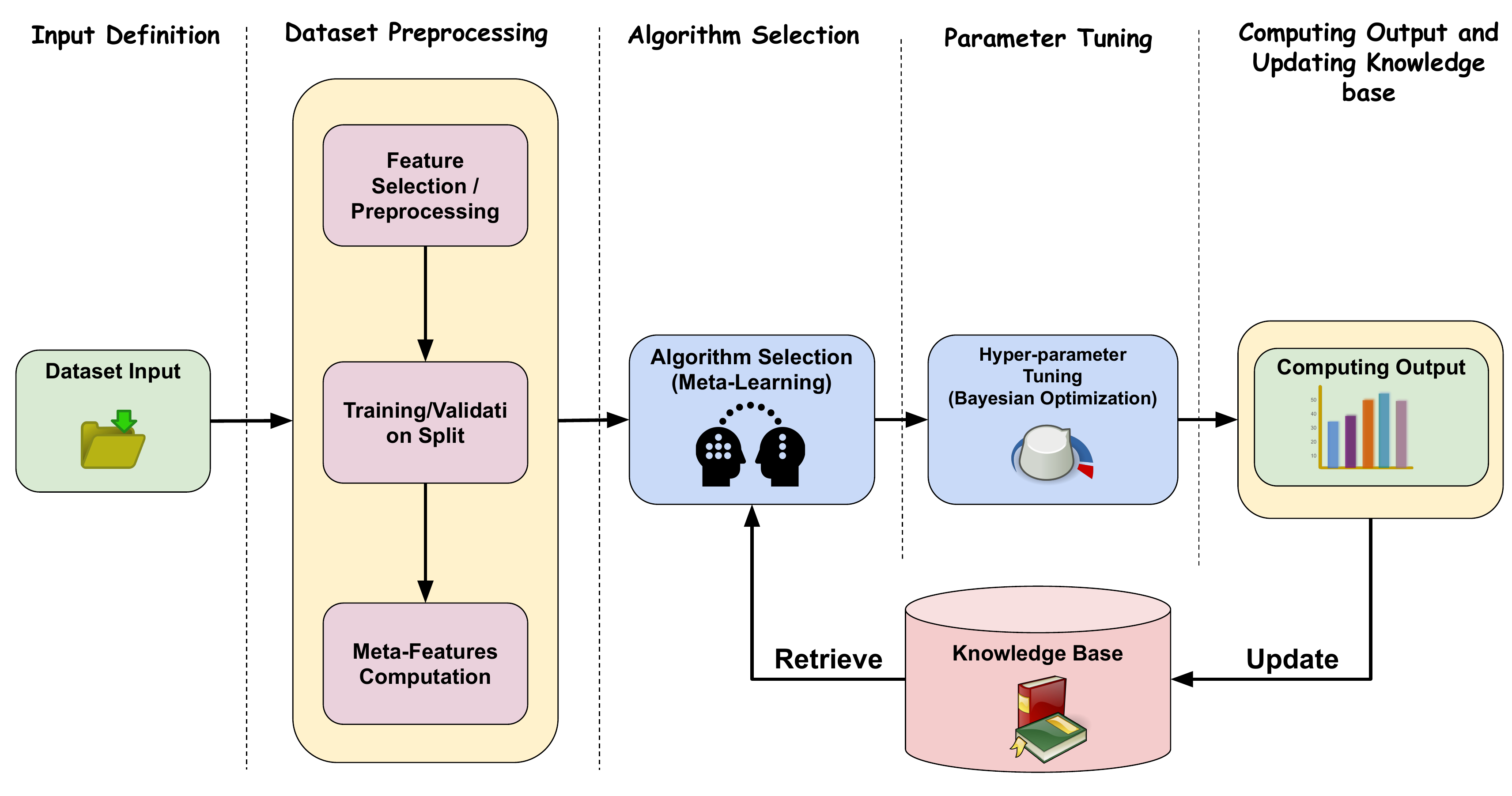}
		\caption{SmartML: Framework Architecture.}
		\label{fig:smartml}
	\end{figure*}

\emph{SmartML}\footnote{\url{https://github.com/DataSystemsGroupUT/SmartML}} has been introduced as the first \texttt{R} package for automated model building for classification tasks~\cite{maher2019smartml}. Figure~\ref{fig:smartml} illustrate the framework architecture of SmartML.
In the algorithm selection phase, SmartML uses a meta-learning approach where the meta-features of the input dataset  is extracted and compared with the meta-features of the datasets that are stored in the framework's knowledge base, populated from the results of the previous runs. The similarity search process is used to identify the similar datasets in the knowledge base, using a nearest neighbor approach, where the retrieved results are used to identify the best performing algorithms on those similar datasets in order to nominate the candidate algorithms for the dataset at hand. The hyper-parameter tuning of SmartML is based on SMAC Bayesian Optimisation~\cite{hutter2011sequential}.  SmartML maintains the results of the new runs to continuously enrich its knowledge base with the aim of further improving the accuracy of the similarity search and thus the performance and robustness for future runs.

\emph{Autostacker}~\cite{Chen:2018:ACE:3205455.3205586} is an AutoML framework that uses an evolutionary algorithm with hierarchical stacking for efficient hyper-parameters search. Autostacker is able to find pipelines, consisting of preprocessing, feature engineering and machine learning algorithms with the best set of hyper-parameters, rather than finding a single machine learning model with the best set of hyper-parameters. Autostacker generates cascaded architectures that allow the components of a pipeline to "correct mistakes made by each other" and hence improves the overall performance of the pipeline. Autostacker simply starts by selecting a set of pipelines randomly. Those pipelines are fed into an evolutionary algorithm that generates the set of winning pipelines.

\emph{AlphaD3M}~\cite{drori2018alphad3m} has been introduced as an AutoML framework that uses meta reinforcement learning to find the most promising pipelines. AlphaD3M finds patterns in the components of the pipelines using recurrent neural networks, specifically long short term memory (LSTM) and Monte-Carlo tree search in an iterative process which is computationally efficient in large search space. In particular, for a given machine learning task over a certain dataset, the network predicts the actions probabilities which lead to sequences that describe the whole pipeline. The predictions of the LSTM neural network are used by Monte-Carlo tree search by running multiple simulations to find the best pipeline sequence.

\emph{OBOE}\footnote{\url{https://github.com/udellgroup/oboe/tree/master/automl}} is an AutoML framework for time constrained model selection and hyper-parameter tuning~\cite{yang2018oboe}. OBOE finds the most promising machine learning model along with the best set of hyper-parameters using collaborative filtering. OBOE starts by constructing an \emph{error matrix} for some base set of machine learning algorithms, where each row represents a dataset and each column represents a machine learning algorithm. Each cell in the matrix represents the performance of a particular machine learning model along with its hyper-parameters on a specific dataset. In addition, OBOE keeps track of the running time of each model on a particular dataset and trains a model to predict the running time of a particular model based on the size and the features of the dataset. Simply, a new dataset is considered as a new row in the error matrix. In order to find the best machine learning algorithm for a new dataset, OBOE runs a particular set of models corresponding to a subset of columns in the error matrix which are predicted to run efficiently on the new dataset. In order to find the rest of the entries in the row, the performance of the models that have not been evaluated are predicted. The good thing about this approach is that it infers the performance of lots of models without the need to run them or even computing meta-features and that is why OBOE can find well performing model within a reasonable time budget.

The \emph{PMF}\footnote{\url{https://github.com/rsheth80/pmf-automl}} AutoML framework  is based on collaborative filtering and Bayesian optimization~\cite{fusi2017probabilistic}. More specifically, the problem of selecting the best performing pipeline for a specific task is modeled as a collaborative filtering problem that is solved using probabilistic matrix factorization techniques. PMF considers two datasets to be similar if they have similar evaluations on a few set of pipelines and hence it is more likely that these datasets will have similar evaluations on the rest of the pipelines. This concept is quite related to collaborative filtering for movie recommendation in which users that had the same preference in the past are more likely to have the same preference in the future. In particular, the PMF framework trains each machine learning pipeline on a sample of each dataset and then evaluates such pipeline. This results in a matrix that summarizes the performance (accuracy or balanced accuracy for classification tasks and RMSE for regression tasks) of each machine learning pipeline of each dataset. The problem of predicting the performance of a particular pipeline on a new dataset can be mapped into a matrix factorization problem.

\emph{VDS}~\cite{alpinemeadow} has been recently introduced as an \emph{interactive} automated machine learning tool, that followed the ideas of a previous work on the~\texttt{MLBase} framework~\cite{kraska2013mlbase}.  In particular, it uses a meta learning mechanism (knowledge from the previous runs) to provide the user  with a quick feedback, in few seconds, with an initial model recommendation  that can achieve a reasonable accuracy while, on the back-end, conducting  an  optimization process so that it can recommend to the user more models with better accuracies, as it progresses with the search process over the search space.
The VDS framework  combines cost-based Multi-Armed Bandits and Bayesian optimizations for exploring the search space while using a rule-based search-space as query optimization technique. VDS prunes unpromising pipelines in early stages using an adaptive pipeline selection algorithm. In addition, it supports wide range of machine learning tasks including classification, regression, community detection, graph matching, image classification, and collaborative filtering. Table~\ref{TBL:CenteralizedFrameWorks} shows a summary of the main features of the centralized state-of-the-art AutoML frameworks.

\begin{sidewaystable*}
\centering
\caption{Summary of the Main Features of Centralized  AutoML Frameworks} {
\scriptsize
\begin{tabular}{|l|c|c|c|c|c|c|c|c|c|}

  \hline

  \textbf{} &
  \begin{tabular}[c]{@{}l@{}}\textbf{Release} \end{tabular} &
  \textbf{Core } & \textbf{Training } & \textbf{Optimization} & \textbf{ML } & \begin{tabular}[c]{@{}l@{}}\textbf{Meta}\end{tabular} & \begin{tabular}[c]{@{}l@{}} \textbf{User}\end{tabular} & \begin{tabular}[c]{@{}l@{}}\textbf{Automatic } \end{tabular}  &
  \begin{tabular}[c]{@{}l@{}}\textbf{Open} \end{tabular} \\

   \textbf{} &
  \begin{tabular}[c]{@{}l@{}}\textbf{} \\ \textbf{Date}\end{tabular} &
  \textbf{Language} & \textbf{Framework} & \textbf{Technique} & \textbf{ Task} & \begin{tabular}[c]{@{}l@{}} \textbf{Learning}\end{tabular} & \begin{tabular}[c]{@{}l@{}} \textbf{Interface}\end{tabular} & \begin{tabular}[c]{@{}l@{}} \textbf{Feature}\end{tabular}  &
  \begin{tabular}[c]{@{}l@{}} \textbf{Source}\end{tabular} \\

   \textbf{} &
  \begin{tabular}[c]{@{}l@{}}\textbf{} \\ \textbf{}\end{tabular} &
  \textbf{} & \textbf{} & \textbf{} & \textbf{ } & \begin{tabular}[c]{@{}l@{}} \textbf{}\end{tabular} & \begin{tabular}[c]{@{}l@{}} \textbf{}\end{tabular} & \begin{tabular}[c]{@{}l@{}} \textbf{Extraction}\end{tabular}  &
  \begin{tabular}[c]{@{}l@{}} \textbf{}\end{tabular} \\

   \hline

  \textbf{AutoWeka} & 2013 & Java & Weka & Bayesian optimization & \begin{tabular}[c]{@{}l@{}} Single-label \\ classification \\ regression \end{tabular} & $\times$ & \checkmark & \checkmark & \checkmark \\
  \hline
  \textbf{AutoSklearn} & 2015 & Python & scikit-learn, & Bayesian optimization & \begin{tabular}[c]{@{}l@{}} Single-label \\ classification \\ regression \end{tabular} & \checkmark &$\times$ & \checkmark & \checkmark \\
  \hline
  \textbf{TPOT} & 2016 & Python & scikit-learn & Genetic Algorithm & \begin{tabular}[c]{@{}l@{}} Single-label \\ classification \\ regression \end{tabular} & $\times$ & $\times$ & \checkmark & \checkmark \\
  \hline
  \textbf{SmartML} & 2019 & R & \begin{tabular}[c]{@{}l@{}} mlr, RWeka \& \\ other R packages \end{tabular} & Bayesian optimization & \begin{tabular}[c]{@{}l@{}} Single-label \\ classification \end{tabular} & \checkmark   & \checkmark & $\times$ & \checkmark \\
  \hline
  \textbf{Auto-MEKA$_{GGP}$} & 2018 & Java & Meka & \begin{tabular}[c]{@{}l@{}} Grammar-based \\ genetic algorithm \end{tabular}  & \begin{tabular}[c]{@{}l@{}} Multi-label \\ classification \end{tabular}  & \checkmark & $\times$  & $\times$ & \checkmark \\
  \hline
   \textbf{Recipe} & 2017 & Python & scikit-learn & \begin{tabular}[c]{@{}l@{}} Grammar-based \\ genetic algorithm \end{tabular}  & \begin{tabular}[c]{@{}l@{}} Single-label \\ classification \end{tabular}  & \checkmark & $\times$  & \checkmark & \checkmark \\
  \hline
  \textbf{MLPlan} & 2018 & Java & Weka and scikit-learn & \begin{tabular}[c]{@{}l@{}} Hierachical\\ Task Planning \end{tabular}  & \begin{tabular}[c]{@{}l@{}} Single-label \\ classification \end{tabular}  & $\times$  & $\times$ & \checkmark & \checkmark \\
  \hline
  \textbf{Hyperopt-sklearn} & 2014 & Python & scikit-learn & \begin{tabular}[c]{@{}l@{}} Bayesian Optimization\\\& Random Search \end{tabular}  & \begin{tabular}[c]{@{}l@{}} Single-label \\ classification \\  regression \end{tabular}  & $\times$ & $\times$ & \checkmark & \checkmark \\
  \hline
  \textbf{Autostacker} & 2018 & - & - & \begin{tabular}[c]{@{}l@{}} Genetic Algorithm \end{tabular}  & \begin{tabular}[c]{@{}l@{}} Single-label \\ classification \end{tabular}  & $\times$ & $\times$ & \checkmark & $\times$\\
  \hline
  \textbf{VDS} & 2019 & - & - & \begin{tabular}[c]{@{}l@{}} cost-based Multi- \\Armed Bandits and \\ Bayesian Optimization \end{tabular}  & \begin{tabular}[c]{@{}l@{}} Single-label \\ classification \\ regression \\ image classification \\ audio classification \\ graph matching  \end{tabular}  & \checkmark & \checkmark & \checkmark & $\times$ \\
  \hline

  \textbf{AlphaD3M} & 2018 & - & - & \begin{tabular}[c]{@{}l@{}} Reinforcement learning \end{tabular}  & \begin{tabular}[c]{@{}l@{}} Single-label \\ classification \\ regression \end{tabular}  & \checkmark & $\times$ & \checkmark & $\times$ \\
  \hline

  \textbf{OBOE} & 2019 & Python & scikit-learn & \begin{tabular}[c]{@{}l@{}}  collaborative filtering \end{tabular}  & \begin{tabular}[c]{@{}l@{}} Single-label \\ classification \end{tabular}  & \checkmark & $\times$ & $\times$ & \checkmark \\
  \hline
   \textbf{PMF} & 2018 & Python & scikit-learn & \begin{tabular}[c]{@{}l@{}} collaborative filtering \& \\ Bayesian
optimization \end{tabular}  & \begin{tabular}[c]{@{}l@{}} Single-label \\ classification \end{tabular}  & \checkmark & $\times$ & \checkmark & \checkmark \\
  \hline

  %\textbf{Deep Learning Models} & Java, Thrift & Java, Scala, & Java, Scala,  & Java, Scala,  & Java, Scala & Java, Python \\
\end{tabular} }
\label{TBL:CenteralizedFrameWorks}
\end{sidewaystable*}

\subsection{Distributed Frameworks}
\begin{figure*}
    \centering
		\includegraphics[width=0.7\linewidth]{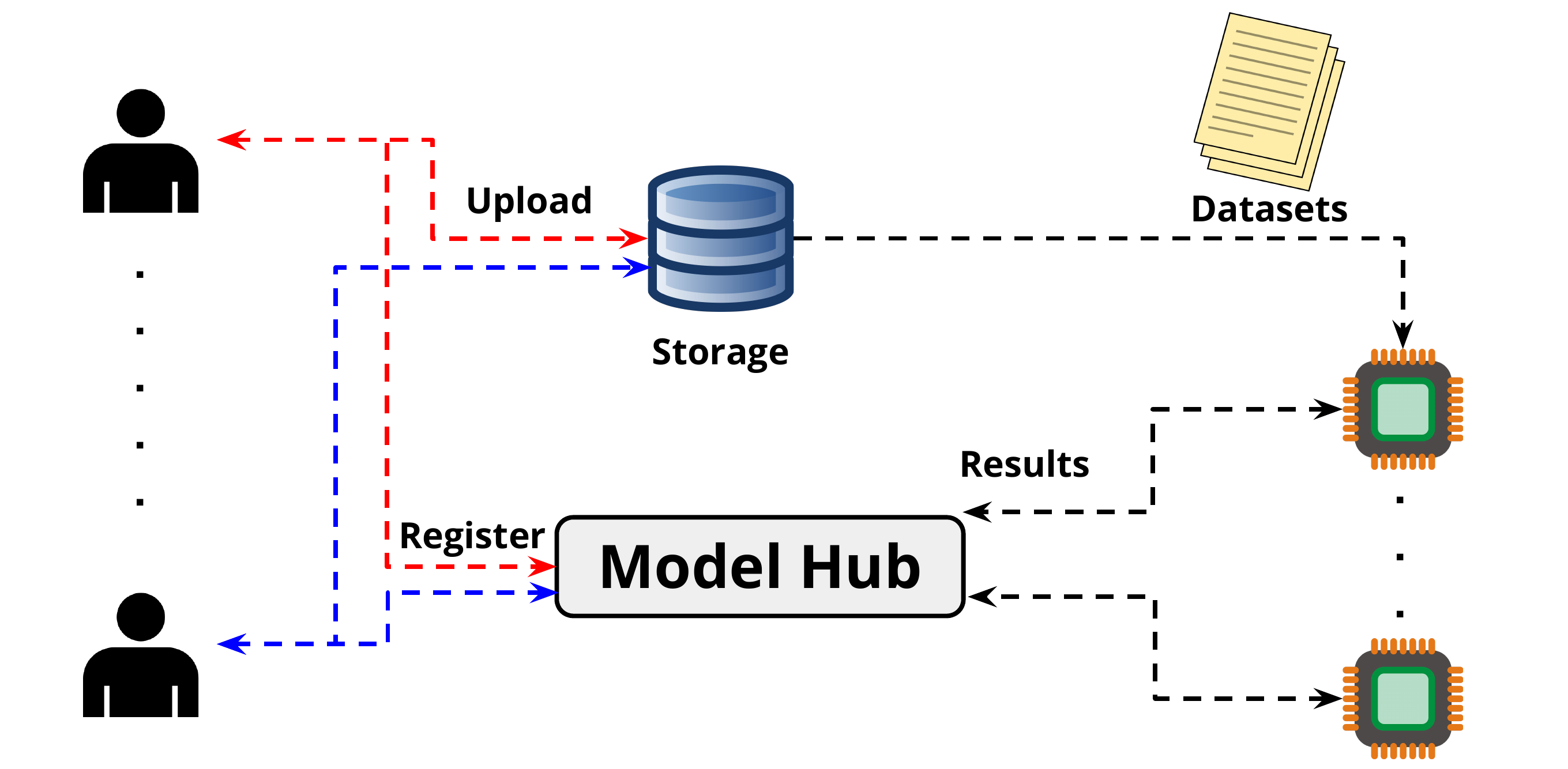}
		\caption{\textbf{ATM: Framework Architecture.}}
		\label{fig:ATM}
	\end{figure*}
As the size of the dataset increases, solving the $CASH$ problem in a centralized
manner turns out to be infeasible due to the limited computing resources (e.g, Memory, CPU) of a single machine. Thus, there is a clear need for distributed solutions that can harness the power of computing clusters that have multiple nodes to tackle the computational complexity of the problem.  \emph{MLbase}\footnote{\url{http://www.mlbase.org/}} has been the first work to introduce the idea of developing a distributed framework of machine learning algorithm selection and hyperparameter optimization~\cite{kraska2013mlbase}. MLbase has been based on \texttt{MLlib}~\cite{meng2016mllib}, a Spark-based ML library. It attempted to reused cost-based query optimization techniques to prune the search space at the level of \emph{logical learning plan} before transforming it into a \emph{physical learning plan} to be executed.

Figure~\ref{fig:ATM} illustrates the architectures of the  \emph{Auto-Tuned Models (ATM)} framework\footnote{\url{https://github.com/HDI-Project/ATM}} that has been introduced as a parallel framework for fast optimization of machine learning modeling pipelines~\cite{inproceedingsATM}. In particular, this framework depends on parallel execution along multiple nodes with a shared model hub that stores the results out of these executions and try to enhance the selection of other pipelines that can out perform the current chosen ones.
The user can decide to use either of ATM's two searching methods, a hybrid Bayesian and multi-armed bandit optimization system, or a model recommendation system that works by exploiting the previous performance of modeling techniques on a variety of datasets.

 \emph{\TransmogrifAI}\footnote{\url{https://transmogrif.ai/}} is one of the most recent modular tools written in Scala. It is built using workflows of feature preprocessors, and model selectors on top of Spark with minimal human involvement. It has the ability to reuse the selected work-flows. Currently, \TransmogrifAI~supports eight different binary classifiers and five regression algorithms. \emph{MLBox}\footnote{\url{https://github.com/AxeldeRomblay/MLBox}} is a  Python-based AutoML framework for distributed preprocessing, optimization and prediction. MLBox supports model stacking where a new model is trained from the combined predictors of multiple previously trained models.
 It uses \texttt{hyperopt}\footnote{\url{https://github.com/hyperopt/hyperopt}}, a distributed asynchronous hyper-parameter optimization library, in Python, to perform the hyper-parameter optimisation process.

 \emph{\Rafiki}\footnote{\url{https://github.com/nginyc/rafiki}}  has been introduced as a distributed framework which is based on the idea of using previous models that achieved high performance on the same tasks~\cite{DBLP:journals/corr/abs-1804-06087}. In this framework, regarding the data and parameters storage, the data uploaded by user to be trained is stored in a Hadoop Distributed File System (HDFS). During training, there is a database for each model storing the best version of parameters from hyper-parameter tuning process. This database is kept in memory as it is accessed and updated
frequently. Once the hyper-parameter tuning process is finished, the database is dumped to the disk. The types of parameters to be tuned are either related to model architecture like number of Layers, and Kernel or related to the training algorithm itself like weight decay, and learning rate. All these parameters can be tuned using random search or Bayesian optimization.
Table~\ref{TBL:DistributedFrameWorks} shows a summary of the main features of the distributed AutoML frameworks.

\begin{table*}
\centering
\caption{Summary of the Main Features of Distributed AutoML Frameworks} {
\scriptsize
\begin{tabular}{|l|c|c|c|c|c|c|c|}
  \hline
  \textbf{} &
  \begin{tabular}[c]{@{}l@{}}\textbf{Release} \\ \textbf{Date}\end{tabular} &
  \begin{tabular}[c]{@{}l@{}}\textbf{Core} \\ \textbf{Language}\end{tabular} &
  \begin{tabular}[c]{@{}l@{}}\textbf{Optimization} \\ \textbf{Technique}\end{tabular} &
  \begin{tabular}[c]{@{}l@{}}\textbf{Training} \\ \textbf{Framework}\end{tabular} &
  \begin{tabular}[c]{@{}l@{}}\textbf{Meta-} \\ \textbf{Learning}\end{tabular} &
  \begin{tabular}[c]{@{}l@{}}\textbf{User} \\ \textbf{Interface}\end{tabular} &
  \begin{tabular}[c]{@{}l@{}}\textbf{Open} \\ \textbf{Source}\end{tabular} \\
   \hline

  \textbf{MLBase} & 2013 & Scala & \begin{tabular}[c]{@{}l@{}} Cost-based Multi- \\Armed Bandits  \end{tabular} & Spark MLlib & $\times$ & $\times$ & $\times$\\
  \hline

  \textbf{ATM} & 2017 & Python & \begin{tabular}[c]{@{}l@{}} Hybrid Bayesian, \\ and Multi-armed \\ bandits Optimization  \end{tabular} & Scikit-Learn & \checkmark & $\times$ & \checkmark\\
  \hline

 \textbf{MLBox} & 2017 & Python & \begin{tabular}[c]{@{}l@{}} Distributed \\ Random search, \\ Tree-Parzen estimators  \end{tabular} & \begin{tabular}[c]{@{}l@{}} Scikit-Learn \\ Keras\end{tabular} & $\times$ & $\times$ & \checkmark \\
  \hline

 \textbf{Rafiki} & 2018 & Python & \begin{tabular}[c]{@{}l@{}} Distributed \\ random search, \\ Bayesian Optimization  \end{tabular} & \begin{tabular}[c]{@{}l@{}} TensorFlow \\ Scikit-Learn\end{tabular} & $\times$ & \checkmark & \checkmark \\
  \hline

  \textbf{TransmogrifAI} & 2018 & Scala & \begin{tabular}[c]{@{}l@{}} Bayesian Optimization, \\ and Random Search \end{tabular} & SparkML & $\times$ & $\times$ & \checkmark \\
  \hline

\end{tabular} }
\label{TBL:DistributedFrameWorks}
\end{table*}

\subsection{Cloud-Based Frameworks}

Several cloud-based solutions have been introduced to tackle the automated machine learning problem using the availability of high computational power on cloud environments to try a wide range of models and configurations.
\emph{Google AutoML}\footnote{\url{https://cloud.google.com/automl/}} has been introduced as a block of the artificial intelligence platform services supported by Google cloud. It supports training a wide range of machine learning models in different domains with minimal user experience. These models can be trained for various tasks including sight, language, and structured data. For instance, AutoML vision, and video intelligence are used in getting insights from visual data like object localization, detection and classification for both static images, and video streams through already pretrained models or training custom models on user data. Similarly, AutoML Natural language, and AutoML translation provide user with APIs for automatic language detection, and transition in addition to insightful text analysis like sentiment classification, and entity extraction. These language services support ten different languages including English, Chinese, French, German and Russian.  On the other hand, AutoML Tables supports training high quality models on tabular structured data by automating feature engineering, model selection, and hyper-parameter tuning steps. Both \emph{Google AutoML} pretrained, and custom models are based on TensorFlow that mainly relies on Google's state-of-the-art transfer learning, neural architecture search technology, and Reinforcement learning with gradient policy upgrade.

\emph{Azure AutoML}\footnote{\url{https://docs.microsoft.com/en-us/azure/machine-learning/service/}} is a cloud-based service that can be used to automate building machine learning pipeline for a both classification and regression tasks. AutoML Azure uses collaborative filtering and Bayesian optimization to search for the most promising pipelines efficiently~\cite{fusi2017probabilistic} based on a database that is constructed by running millions of experiments of evaluation of different pipelines on many datasets. This database helps in finding the good solutions for new datasets quickly. \emph{Azure AutoML} is available in the Python SDK of Microsoft Azure machine learning and it is based on scikit-learn search space of different learning algorithms. In addition, it gives the user the flexibility to use this service either locally or leveraging the performance and scalability of Azure cloud services.

\emph{Amazon Sage Maker}\footnote{\url{https://aws.amazon.com/machine-learning/}} provides its users with a wide set of most popular machine learning, and deep learning frameworks to build their models in addition to automatic tuning for the model parameters. \emph{Sage Maker} supports automatic deployment for models on auto-scaling clusters in multiple zones to ensure the high availability, and performance during generation of predictions. Moreover, Amazon offers a long list of pretrained models for different AI services that can be easily integrated to user applications including different image and video analysis, voice recognition, text analytics, forecasting, and recommendation systems.

\subsection{Neural Network Automation Frameworks}
 Recently, some frameworks (e.g., \texttt{Auto-Keras}~\cite{DBLP:journals/corr/abs-1806-10282}, and \texttt{Auto-Net}~\cite{mendoza2016towards}) have been proposed with the aim of automatically finding neural network architectures that are competitive with architectures designed by human experts. However, the results so far are not significant. For example, \emph{Auto-Keras}~\cite{DBLP:journals/corr/abs-1806-10282} is an open source efficient neural architecture search framework based on Bayesian optimization to guide the network morphism. In order to explore the search space efficiently, Auto-Keras uses a neural network kernel and tree structured acquisition function with iterative Bayesian optimization. First, a Gaussian process model is trained on the currently existing network architectures and their performance is recorded. Then, the next neural network architecture obtained by the acquisition function is generated and evaluated. Moreover, Auto-Keras runs in a parallel mode on both CPU and GPU.

\begin{table*}
\centering
\caption{Summary of the Main Features of the n]Neural Architecture Search frameworks} {
\scriptsize
\begin{tabular}{|l|c|c|c|c|c|c|c|}
  \hline
  \textbf{} & \begin{tabular}[c]{@{}l@{}}\textbf{Release} \\ \textbf{Date}\end{tabular}  & \textbf{Open Source} & \textbf{Optimization technique} & \textbf{Supported Frameworks} & \textbf{Interface} \\
   \hline

  \textbf{Auto Keras} & 2018 & \checkmark & \begin{tabular}[c]{@{}l@{}} Network Morphism \end{tabular} & Keras & \checkmark\\
  \hline
 \textbf{Auto Net} & 2016 & \checkmark & SMAC & PyTorch & $\times$\\
  \hline
  \textbf{NNI} & 2019 & \checkmark & \begin{tabular}[c]{@{}l@{}} Random Search \\ Different Bayesian Optimizations \\ Annealing \\ Network Morphism \\ Hyper-Band \\ Naive Evolution \\ Grid Search  \end{tabular} & \begin{tabular}[c]{@{}l@{}} PyTorch, \\ TensorFlow, \\ Keras, \\ Caffe2, \\CNTK, \\Chainer \\ Theano \end{tabular} & \checkmark \\
  \hline
  \textbf{enas} & 2018 & \checkmark & \begin{tabular}[c]{@{}l@{}} Reinforcement Learning  \end{tabular} & Tensorflow & $\times$ \\
  \hline
  \textbf{NAO} & 2018 & \checkmark & \begin{tabular}[c]{@{}l@{}} Gradient based optimization  \end{tabular} & Tensorflow, PyTorch & $\times$ \\
  \hline
  \textbf{DARTS} & 2019 & \checkmark & \begin{tabular}[c]{@{}l@{}}Gradient based optimization \end{tabular} & PyTorch & $\times$ \\
  \hline
  \textbf{LEAF} & 2019 & $\times$ & \begin{tabular}[c]{@{}l@{}}Evolutionary Algorithms \end{tabular} & - & $\times$ \\
  \hline

\end{tabular} }
\label{TBL:DeepLearningFrameWorks}
\end{table*}

\emph{Auto-Net}~\cite{mendoza2016towards} is an efficient neural architecture search framework based on SMAC optimization and built on top of \texttt{PyTorch}. The first version of Auto-Net is implemented within the Auto-sklearn in order to leverage some of the existing components of the of the machine learning pipeline in Auto-sklearn such as preprocessing. The first version of Auto Net only considers fully-connected feed-forward neural networks as they are applied on a large number of different datasets. Auto-net accesses deep learning techniques from Lasagne Python deep learning library~\cite{dieleman2016lasagne}. Auto Net includes a number of algorithms for tuning the neural network weights including vanilla stochastic gradient descent , stochastic gradient descent with momentum, Adadelta~\cite{zeiler2012adadelta}, Adam~\cite{kingma2014adam}, Nesterov momentum~\cite{nesterov27method}and Adagrad~\cite{duchi2011adaptive}.

\emph{Neural Network Intelligence}(NNI)\footnote{\url{https://github.com/Microsoft/nni}} is an open source toolkit by Microsoft that is used for tuning neural networks architecture and hyper-parameters in different environments including local machine, cloud and remote servers. NNI accelerates and simplifies the huge search space using built-in super-parameter selection algorithms including random search, naive evolutionary algorithms, simulated annealing, network morphism, grid search, hyper-band, and a bunch of Bayesian optimizations like SMAC~\cite{hutter2011sequential}, and BOHB~\cite{falkner2018bohb}. NNI supports a large number of deep leaning frameworks including \texttt{PyTorch}, \texttt{TensorFlow}, \texttt{Keras}, \texttt{Caffe2}, \texttt{CNTK}, \texttt{Chainer} and \texttt{Theano}.

\emph{DEvol}~\footnote{\url{https://github.com/joeddav/devol}} is an open source framework for neural network architecture search that is based on genetic programming to evolve the number of layers, kernels, and filters, the activation function and dropout rate. DEvol uses parallel training in which multiple members of the population are evaluated across multiple GPU machines in order to accelerate the process of finding the most promising network.

\emph{enas}~\cite{pham2018efficient} has been introduced as an open source framework for neural architecture search in Tensorflow based on reinforcement learning~\cite{zoph2016neural} where a controller of a recurrent neural network architecture is trained to search for optimal subgraphs from large computational graphs using policy gradient. Moreover, \emph{enas} showed a large speed up in terms of GPU hours thanks to the sharing of parameters across child subgraphs during the search process.

\emph{NAO}~\cite{luo2018neural}, and \emph{Darts}~\cite{liu2018darts} are open source frameworks for neural architecture search which propose a new continuous optimization algorithm that deals with the network architecture as a continuous space instead of the discretization followed by other approaches. In \emph{NAO}, the search process starts by encoding an initial architecture to a continuous space. Then, a performance predictor based on gradient based optimization searches for a better architecture that is decoded at the end by a complementary algorithm to the encoder in order to map the continuous space found back into its architecture. On the other hand, \emph{DARTS} learns new architectures with complex graph topologies from the rich continuous search space using a novel bilevel optimization algorithm. In addition,
it can be applied to any specific architecture family without restrictions to any of convolutional and recurrent networks only. Both frameworks showed a competitive performance using limited computational resources compared with other neural architecture search frameworks.

\emph{Evolutionary Neural AutoML for Deep Learning (LEAF)}~\cite{liang2019evolutionary} is an AutoML framework that optimizes neural network architecture and hyper-parameters using the state-of-the-art evolutionary algorithm and distributed computing framework. LEAF uses \texttt{CoDeepNEAT}~\cite{miikkulainen2019evolving} for optimizing deep neural network architecture and hyper-parameters. LEAF consists of three main layers which are algorithm layers, system layer and problem-domain layer. LEAF evolves deep neural networks architecture and hyper-parameters in the algorithm layer. The system layer is responsible for training the deep neural networks in a parallel mode on a cloud environment such as Microsoft Azure\footnote{\url{https://azure.microsoft.com/en-us/}}, Google Cloud\footnote{\url{https://cloud.google.com/}} and Amazon AWS\footnote{\url{https://aws.amazon.com/}}, which is essential in the evaluation of the fitness of the neural networks evolved in the algorithm layer. More specifically, the algorithm layer sends the neural network architecture to the system layer. Then, the system layer sends the evaluation of the fineness of this network back to the algorithm layer. Both the algorithm layer and the system layer work together to support the problem-domain layers where the problems of hyper-parameter tuning of  network architecture search are solved.
Table~\ref{TBL:DeepLearningFrameWorks} shows summary of the main features of the state-of-the-art neural architecture search frameworks.

\section{Other Automation Aspects In \\ Model  Building Life Cycle}
\label{SEC:Other}

\begin{figure*}
		\includegraphics[width=\linewidth]{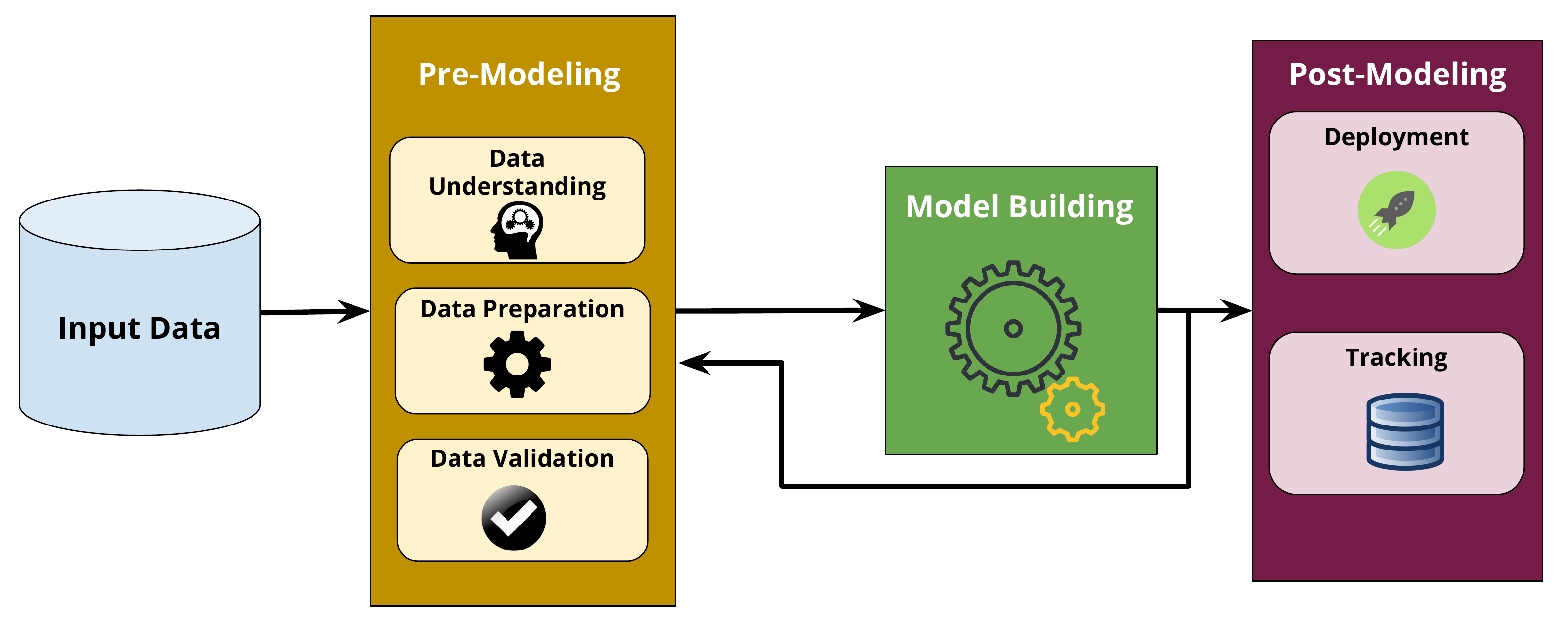}
		\caption{Machine Learning Production Pipeline}
		\label{fig:mlPipeline2}
	\end{figure*}

\begin{figure*}[t]
    \centering
		\includegraphics[width=0.75\linewidth]{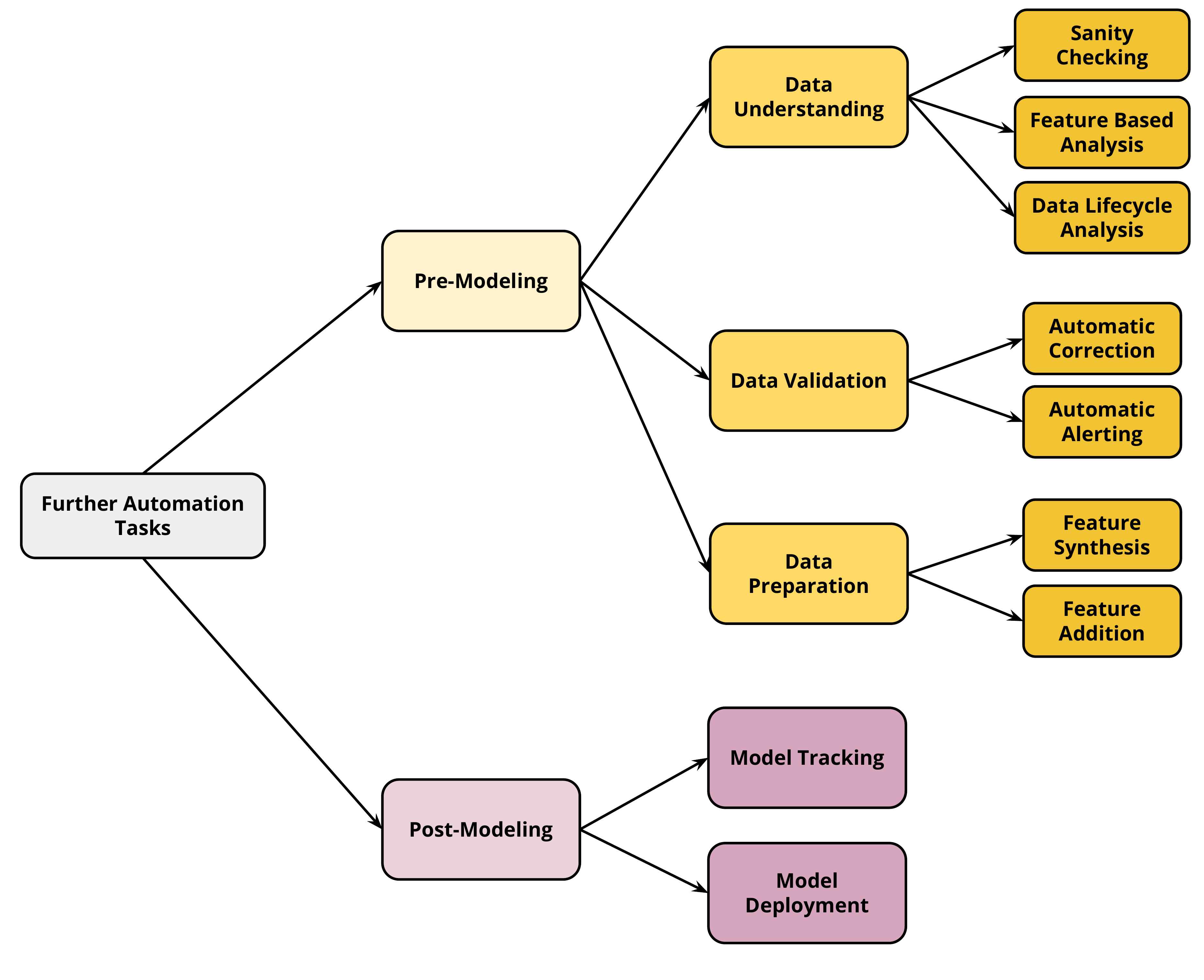}
		\caption{ \textbf{Further automation tasks that ease the data scientist work.}}
		\label{fig:automlTaxonomy}
	\end{figure*}

While current different AutoML tools and frameworks have minimized the role of data scientist in the modeling part and saved much effort, there are still several aspects that need human intervention and interpretability in order to make the correct decisions that can enhance and affect the modeling steps. These aspects belongs to two main building blocks of the machine learning production pipeline: \emph{Pre-Modeling} and \emph{Post-Modeling} (Figure~\ref{fig:mlPipeline2}). In general, \emph{Pre-Modeling} is an important block of the  machine learning pipeline that can dramatically affect the outcomes of the automated algorithm selection and hyper-parameters optimization process. The pre-modeling step includes a number of steps including data understanding, data preparation and data validation. In addition, the \emph{Post-Modeling} block covers other important aspects including the management and deployment of produced machine learning model which represents  a corner stone in the pipeline that requires the ability of packaging model for reproducibility. The aspects of these two building blocks can help on covering what is missed in current AutoML tools, and help data scientists in doing their job in a much easier, organized, and informative way. In this section, we give an overview of a number of systems and frameworks that have been developed to aid the data scientists on the steps of the pre-modeling and post-modeling blocks (Figure~\ref{fig:automlTaxonomy}).

\subsection{Pre-Modeling}
    \subsubsection{Data Understanding}
    The data understanding step mainly focuses on formulating alerts that can be easily solved by identifying suitable actions. The key point of data understanding is the sensitivity of alerts to data errors. Available tools try to automate the process of data understanding and reduce human involvement in this stage. However, it is still essential to have a human to confirm the actions taken. There are three possible scenarios for data understanding:

    	\textbf{Sanity Checking}:
    	The sanity checking  process is used to ensure that the feature is suitable for being used to train a model. For example, a feature that has 70\% missing values would not be a good one for training the model or the negative age values should be corrected. \seeDB~\cite{journals/pvldb/VartakRMPP15} has been introduced as a visualization recommendation engine that eases the analysis of data subsets in an efficient way by showing a large number of recommended visualizations in addition to choosing the appropriate metrics to measure the visualization interest in a near interactive performance with a reduced latency by over 100X than traditional techniques. \seeDB~can be run on top of a wide range of database management systems. \zenVisage~\cite{Siddiqui:2016:EDE:3025111.3025126} extends the prior system \seeDB. In addition to the visualization recommendation, it supports a graph based visual query language called ZQL which is a flexible technique to choose the desired insights from visualizations using very small lines. Moreover, \zenVisage~provides the user with an interactive interface which enables him to directly draw different chart types or specify the trend of current interest to be visualized. \QUDE~\cite{Zhao:2017:CFD:3035918.3064019} presents a solution using a novel $\alpha$-investing technique to the multiple hypothesis testing error introduced by previous tools which is making several false discoveries based on inference seen by users from simple interactions visualized by these tools. \QUDE has showed a significant decrease in the false discovery rate than other interactive tools for both synthetic and real-world datasets.
    	
    	\textbf{Feature Based Analysis}:
    	Data understanding is not limited only to performing a sanity check before model building. It is an iterative process that can occur even after model building to identify the data slices that affect the model quality and behavior. \MLCube~\cite{Kahng:2016:VEM:2939502.2939503} is a popular framework for the feature-based analysis of data understanding. It computes several evaluation metrics and statistics over subsets of data defined by some feature conditions on a machine learning model. \MLCube~integrates its results with an interactive explorer for visualization and models performance comparison. \texttt{Smart Drill-Down}~\cite{Joglekar:2015:SDN:2824032.2824103} has been introduced as an operator that is used to explore, investigate and summarizes group of rows in a relational table. It presents a solution to present the interesting aspects in parts of a table that can affect the model behavior. Smart Drill-Down uses an approximation algorithm to find the optimal list of interesting rules as it is a NP-Hard problem.
    	
    	\textbf{Data Life-cycle Analysis}:
    	Identifying the sources of data errors from a given model and interpreting the dependencies between features of dataset is an important part of data understanding through the whole data life cycle. \GOODs~\cite{Halevy:2016:GOG:2882903.2903730} is a framework that has been introduced to organize structured datasets of different formats at a scale. It extracts meta-data from each dataset such as schema, and timestamps to draw the similarity and provenance relations between these datasets. It allows users to find similar datasets, monitor, and annotate them. \ProvDB~\cite{DBLP:journals/corr/MiaoCD16} is a prototype system that has been built on top of \texttt{git} and \texttt{Neo4j}~\cite{Kemper:2015:BN:2888529} graph database with the aim of keeping track of the analyses performed and datasets generated as a meta-data management system that enables collaborative work flows. \ProvDB~allows data scientists to query the captured information. Also, it can help in identifying flaws that can be extracted from data science process itself in addition to automatic monitor, and analysis of deployed models. Ground~\cite{DBLP:conf/cidr/HellersteinSGSA17} is an open-source data context service that manages the data storage which facilitates the good use of data. This system was named as \Ground~to indicate their target of unifying the ground of data scientists by integrating some underground services like crawlers, version storage, search index, scheduling, and work flow. Moreover, there are some above ground applications like model serving, reproducibility, analytics and visualization tools, information extractors and security auditing. This unified ground will help to create a data context layer in the big data stack. It is a community effort that can provide useful open source where several applications, and services can be easily integrated and plugged into the same ground.

    \subsubsection{Data Validation}
    Data validation is the block that separates the data preparation from model training in the machine learning production pipeline. Several aspects are considered in this step such as adding new features, cleaning the existing ones before building or updating the model, and automatically inserting corrections to invalid data.

        \textbf{Automatic Data invalidity Diagnosis and correction}:
        In general, problems may occur in datasets, especially incremental ones, that affect  its coherency. For example, some data labels go from capital to lower case, some feature has various currency values, or person age written in different formats as number of years or months. This data invalidity requires automatic fixation at the time of insertion. \DataXRay~\cite{Wang:2015:DXD:2723372.2750549} has been introduced as a  diagnostic tool for data systematic errors that are inherited from the process of producing the data itself. It is implemented over concept of MapReduce to allow its scalability. The main contributions of this tool are designing a fast algorithm for diagnosing data on a large-scale, using Bayesian analysis to formulate a model that defines the good diagnosis principles, and transforming the normal diagnosis problem into a search task for finding the common properties of the erroneous features.
        Experiments made on this tool outperforms the alternative techniques like feature selection algorithms and showed that it can effectively identify the causes of data errors. \MacroBase \cite{Bailis:2017:MPA:3035918.3035928} is an open source framework for data analytics, and search engine for big, and fast data streams. It enables modular, fast and accurate analysis that can detect and monitor unusual behavior in data and deliver summaries of important landmarks over the data streams that represent the unusual behavior. \MacroBase~achieves a speed up to 2 million events per second for each query on one core. The main contribution of \MacroBase~is its ability to optimize the combination of explanation and classification tasks for fast data streams.

        \textbf{Alert Combining}:
        When there are multiple alerts, the system should be able to relate them together and determine the root cause of these alerts to facilitate the process of automatic repair by combining multiple alerts into a few ones. \cite{Bohannon:2005:CME:1066157.1066175} proposes a framework that applies different techniques from record-linkage to the search of good repairs by introducing an approach defined with two greedy algorithms with a cubic time complexity in the database size.
        In addition, some optimizations are added if there is any presence of duplicate instances or records that greatly improve the performance and scalability. Experimental results on this framework showed a great improvement in the performance with little cost of the repair quality. \texttt{Conflict Hypergraph}~\cite{Kolahi:2009:AOR:1514894.1514901} has been introduced as an approximation algorithm which can be used to clean, and repair inconsistencies of a fixed set of functional dependencies
        with the minimum number of modifications in databases. This algorithm tries to find a solution that is far from the optimum repair which is NP-Hard, with distance less than a fixed constant
        factor.

    \subsubsection{Data Preparation}
        Data preparation is considered as the most time consuming stage in the pipeline due to the presence of many
        various data preprocessing algorithms including Normalization, Bucketization, Winsorizing, One-Hot encoding,
        Feature Crosses, etc.  In addition, new features can be synthesized from current available data that are better representatives for patterns in data and eases the role of the modeling process. Solutions in the literature either merge the feature preprocessing algorithms with the model training phase as a top layer over algorithm selection that needs to be optimized too. However, some other solutions depends on different types of auto encoder, Restricted Boltzmann Machines (RBM),~\cite{Fischer:2014:TRB:2533326.2533561} by feeding the data features directly to a deep neural networks that can figure out the best features which can be generated using stochastic artificial neural networks that can learn the probability distribution over the set of data inputs.

        Another possibility for model performance improvement is adding more features to the dataset. However, finding suitable and complementary data is a very difficult task. Thus, many trials have been made to create a repositories for the wide range of datasets in different domains. Recently, Google has initiated a search engine for datasets\footnote{Google Dataset Search \url{https://toolbox.google.com/datasetsearch}}. In addition, \openML \cite{OpenML2013} is a well-organized data repository that allow users to find different datasets, data analysis flows, explore results of these flows, and create tasks that could be shared with the data scientists community and ease the process of finding a solution. Furthermore, in practice, there is a need for version control system for datasets which is offered by \DataHub~\cite{DBLP:journals/corr/BhardwajBCDEMP14} that can keep track of different datasets versions, and allows collaborative incremental work on data or roll backs in case of error occurrences.
        In particular, \Datahub~is a web client for dataset version control system like git, used for easy manipulation of datasets. It provides a complete ecosystem data ingestion, processing, repairing, integration, discovery, query, visualization, and analytics with a Restful API. It allows a user to share data, collaborate with others, and perform a wide range of operations from the available suite of tools to process the data. In addition, there several attempts for automatic feature synthesis from available data features. For instance, \emph{Feature Tools}~\cite{kanter2015deep} is a library for automated feature engineering which follows a deep feature synthesis algorithm that can work with relational databases making use of the entity, forward, and backward relations between tables in generating new higher level features iteratively that can improve the modeling performance.

\subsection{Post-Modeling}
In practice, there is an urgent need to try the integration of best algorithms, and tools in different pipeline phases in a single
workflow. This step will be the corner-stone of data scientist replacement. Recently, \mlFlow~has been introduced as  an open source platform to manage the machine learning pipeline from end-to-end. It is
a language agnostic platform that has a REST API, and Command-Line interface in addition to APIs for most popular programming languages like Python, R, and Java. The \mlFlow~ performs three different operations which are:
\begin{itemize}
	\item Record results from experiments and work flows made by different tools and algorithms. In addition, code versions with metrics used, parameter configurations, and visualizations made can all be tracked and stored.
	\item Package the code used in a chain of reusable and reproducible format to be shared with all the community or to be transferred directly to production. Over and above, it handles all the needed dependencies, and entry points.
	\item Manage and Deploy the models built from the different work flows over wide range of platforms.
\end{itemize}
\mlFlow~ facilitates many tasks for the data scientist. However, it still lacks the smartness of recommending best work flows that are suitable for each task, and requires human interaction in taking several actions and solving conflicts that occur by transferring models between different platforms.

Similarly, \emph{ModelChimp}\footnote{\url{https://modelchimp.com/}}~provides a server based solution for tracking machine learning, and deep learning experiments that can be connected to an external PostgreSQL database for easily storage and retrieval of results. In addition, it supports real-time visualization for tracking the training process with different metrics, and parameters. \emph{ModelChimp} supports most popular frameworks such as \texttt{scikit-learn}, \texttt{Keras}, \texttt{PyTorch}, and \texttt{TensorFlow}. Additionally, \emph{datmo}\footnote{\url{https://github.com/datmo/datmo}}~is an open source tool, in Python, for production model management that helps data scientists to store experiments logs and results with easy reproducibility, and project versioning. Moreover, it allows synchronization between these stored logs with user private cloud storage folders.

\section{Open Challenges and Future Directions}
\label{SEC:open}
Although in the last years, there has been increasing research efforts to tackle the challenges of the automated machine learning domain, however, there are still several open challenges and research directions that needs to be tackled to achieve the ultimate goals and vision of the AutoML domain. In this section, we highlight some of these challenges that need to be tackled to improve the state-of-the-art.

 \textbf{Scalability}: In practice, a main limitation of the centralized frameworks for automating the solutions for the CASH problem (e.g., \texttt{Auto-Weka}, \texttt{Auto-Sklearn}) is that they are tightly coupled with a machine learning library (e.g., \texttt{Weka}, \texttt{scikit-learn}, \texttt{R}) that can only work on a \emph{single} node which makes them not applicable in the case of large data volumes. In practice, as the scale of data produced daily is increasing continuously at an exponential scale, several distributed machine learning platforms have been recently introduced. Examples include \texttt{Spark MLib}~\cite{meng2016mllib}, \texttt{Mahout}\footnote{\url{https://mahout.apache.org/}} and \texttt{SystemML}~\cite{boehm2016systemml}. Although there have been some initial efforts for distributed automated framework for the CASH problem. However, the proposed distributed solutions are still simple and limited in their capabilities. More research efforts and novel solutions are required to tackle the challenge of automatically building and tuning machine learning models over massive datasets.

 \textbf{Optimization Techniques}: In practice, different AutoML frameworks use different techniques for hyper-parameter optimization of the machine learning algorithms. For instance, \texttt{Auto-Weka} and \texttt{Auto-Sklearn} use the SMAC technique  with cross-fold validation during the hyper-parameter configuration optimization and evaluation. On the other hand, \texttt{ML-Plan} uses the hierarchical task network with Monte Carlo Cross-Validation. Other tools, including \texttt{Recipe}~\cite{de2017recipe} and \texttt{TPOT}, use genetic programming, and pareto optimization for generating candidate pipelines. In practice, it is difficult to find a clear winner or one-size-fits-all technique. In other words, there is no single method that will be able to outperform all other techniques on the different datasets with their various characteristics, types of search spaces and metrics (e.g., time and accuracy). Thus, there is a crucial need to understand the Pros and Cons of these optimization techniques so that  AutoML systems can automatically tune their hyper-parameter optimization techniques or their strategy for exploring and traversing the search space. Such decision automation should provide improved performance over picking and relying on a fixed strategy. Similarly, for the various introduced meta-learning techniques, there is no clear systematic process or evaluation metrics to quantitatively assess and compare the impact of these techniques on reducing the search space. Recently, some competitions and challenges\footnote{\url{https://www.4paradigm.com/competition/nips2018}}$^,$\footnote{\url{http://automl.chalearn.org/}}  have been introduced and organized to address this issue such as the DARPA D3M Automatic Machine Learning competition~\cite{alpinemeadow}.

 \textbf{Time Budget}: A common important parameter for AutoML systems is the user  \emph{time budget} to wait before getting the recommended pipeline. Clearly, the bigger the time budget, the more the chance for the AutoML system to explore various options in the search space and the higher probability to get a better recommendation. However, the bigger time budget used, the longer waiting time and the higher computing resource consumption, which could be translated into a higher monetary bill in the case of using cloud-based resources. On the other hand, a small-time budget means a shorter waiting time but a lower chance to get the best recommendation. However, it should be noted that increasing the time budget from $X$ to $2X$ does not necessarily lead to a big increase on the quality of the results of the recommended pipeline, if any at all. In many scenarios, this extra time budget can be used for exploring more of the unpromising branches in the search space or exploring branches that have very little gain, if any.
 For example, the accuracy of the returned models from running the \texttt{AutoSklearn} framework over the \texttt{Abalone} dataset\footnote{\url{https://www.openml.org/d/183}} with time budgets of 4 hours and 8 hours are almost the same (25\%).
 Thus, accurately estimating or determining the adequate time budget to optimize this trade-off is  another challenging decision that can not be done by non-expert end users. Therefore, it is crucial to tackle such challenge by automatically predicting/recommending the adequate time budget for the modeling process. The \texttt{VDS}~\cite{alpinemeadow} framework provided a first attempt to tackle this challenge by proposing an interactive approach that relies on meta learning to provide a quick first model recommendation that can achieve a reasonable quality while conducting an offline optimization process and providing the user with a \emph{stream} of models with better accuracy. However, more research efforts to tackle this challenge are still required.

\textbf{Composability}
Nowadays, several machine learning solutions (e.g., \texttt{Weka}, \texttt{Scikit-Learn}, \texttt{R}, \texttt{MLib}, \texttt{Mahout}) have become popular. However, these ML solutions significantly vary in their available techniques (e.g., learning algorithms, preprocessors, and feature selectors) to support each phase of the machine learning pipeline. Clearly, the quality of the machine learning pipelines that can be produced by any of these platforms depends on the availability of several techniques/algorithms that can be utilized in each step of the pipeline. In particular, the more available techniques/algorithms in a machine learning platform, the higher the ability and probability of producing a well-performing machine learning pipeline. In practice, it is very challenging to have optimized implementations for all of the algorithms/techniques of the different steps of the machine learning pipeline available in a single package, or library.
The \texttt{ML-Plan} framework~\cite{DBLP:journals/ml/MohrWH18} has been attempting to tackle the composability challenge on building machine learning pipelines. In particular, it integrates a superset of both \texttt{Weka} and \texttt{Scikit-Learn} algorithms to construct a full pipeline. The initial results of this approach have shown that the composable pipelines over \texttt{Weka} and \texttt{Scikit-Learn} do not significantly outperform the outcomes from \texttt{Auto-Weka} and \texttt{Auto-Sklearn} frameworks especially with big datasets and small time budgets. However, we believe that there are several reasons behind these results. First, combining the algorithms/techniques of more than one machine learning platform causes a dramatic increase in the search space. Thus, to tackle this challenge, there is a crucial need for a smart and efficient search algorithm that can effectively reduce the search space and focus on the promising branches. Using meta-learning approaches can be an effective solution to tackle this challenge. Second, combining services from more than one framework can involve a significant overhead for the data and message communications between the different frameworks. Therefore, there is a crucial need for a smart \emph{cost-based} optimizer that can accurately estimate the gain and cost of each recommended composed pipeline and be able to choose the composable recommendations when they are able to achieve a clear performance gain. Third, the \texttt{ML-Plan} has been combining the services of two single node machine learning services (\texttt{Weka} and \texttt{Scikit-Learn}). We believe that the best gain of the composability mechanism will be achieved by combining the performance power of distributed systems (e.g., \texttt{MLib})  with the rich functionality of many centralized systems.

 \textbf{User friendliness}: In general, most of the current tools and framework can not  be considered to be user friendly. They still need sophisticated technical skills to be deployed and used. Such challenge limits its usability and wide acceptance among layman users and domain experts (e.g., physicians, accountants) who commonly have limited technical skills. Providing an interactive and light-weight web interfaces for such framework can be one of the approaches to tackle these challenges.

 \textbf{Continuous delivery pipeline}: Continuous delivery is defined as creating a repeatable, reliable and incrementally improving process for taking software from concept to customer. Integrating machine learning models into continuous delivery pipelines for productive use has not recently drawn much attention, because usually the data scientists push them directly into the production environment with all the drawbacks this approach may have, such as no proper unit and integration testing.

 \textbf{Data Validation}: In this context, most of the solutions in literature focus on problem detection and user notification only. However, automatic correction hasn't been investigated in a good manner that covers several possible domains of datasets and reduce the data scientist's role in machine learning production pipeline. In addition, as the possible data repairing is a NP-Hard problem,  there is a need to find more approximation techniques that can solve this problem.

\textbf{Data Preparation}: In practice, there is a crucial need for automating the feature extraction process as it is considered as one of the most time consuming part of the pipeline. In practice, most of the systems neglect the automation of transferring data features into different domain space like performing principal component analysis, or linear discriminant analysis and when they improve the model performance. In addition, we believe that there is a room for improvement of current auto-encoders types like restricted Boltzmann Machines. So, further research is needed to try different architectures and interpret them to have the ability to automate the choice of suitable encoders. Furthermore, there are various techniques for measuring a score for the feature importance which is a very important part to automate the feature selection process. However, there is no comprehensive comparative studies between these methods or good recipes that can recommend when to use each of these techniques.

\textbf{Model Deployment and Life Cycle}: Recently, there some tools and frameworks that have been introduced to ease the data scientist work and automate the machine learning production. However, in practice,  there is still a need to integrate these different systems along the whole pipeline. For example, there is a large room for improvement regarding the automatic choice of the good work flows specific to each problem and how to integrate more data understanding, validation and preparation techniques with the work flows. In particular, these frameworks are still not providing the end-user with any \emph{smartness} in the decision making process which is a corner stone towards replacing the role of human in the loop.

\section{Conclusion}
\label{SEC:Conclusion}
Machine learning has become one of the main engines of the current era. The production pipeline of a machine learning models passe through different phases and stages that require wide knowledge of several available tools, and algorithms. However, as the scale of data produced daily is increasing continuously at an exponential scale, it has become essential to automate this process. In this survey, we have covered comprehensively the state-of-the-art research effort in the domain of AutoML frameworks. We have also highlighted research directions and open challenges that need to be addressed in order to achieve the vision and goals of the AutoML process. We hope that our survey serves as a useful resource for the community, for both researchers and practitioners, to understand the challenges of the domain and provide useful insight for further advancing the state-of-the-art in several directions.

\section*{Acknowledgment}
This work of Sherif Sakr and Mohamed Maher is funded by the European Regional Development Funds via the Mobilitas Plus programme
(grant MOBTT75).
The work of Radwa Elshawi is funded by the European Regional Development Funds via the Mobilitas Plus programme (MOBJD341).

% Bibliography
\bibliographystyle{plain}
\bibliography{Biblio}

% History dates
%\received{October 2018}{October 2018}{October 2018}
\end{document}